\documentclass[10pt,twocolumn,letterpaper]{article}
\usepackage[pagenumbers]{cvpr}

\usepackage{multirow}
\usepackage{tabularx}
\usepackage{colortbl}
\usepackage[utf8]{inputenc}
\usepackage[T1]{fontenc}
\usepackage[table]{xcolor}
\usepackage{graphicx}
\usepackage{booktabs}
\usepackage{amssymb}
\usepackage{enumitem}
\usepackage{array}
\usepackage{makecell}
\usepackage{wrapfig}
\usepackage{url}
\usepackage{pifont}
\newcommand{\cmark}{\ding{51}}
\newcommand{\xmark}{\textnormal{\sffamily\mdseries\upshape X}}
\newcolumntype{Y}{>{\centering\arraybackslash}X}

\usepackage{placeins}
\newsavebox{\stalTableMeasure}
\newcommand{\stalFullWidthTable}[3]{%
  \begingroup
  \sbox{\stalTableMeasure}{\begin{tabular}{#2}#3\end{tabular}}%
  \setlength{\tabcolsep}{\dimexpr\tabcolsep+(\linewidth-\wd\stalTableMeasure)/#1\relax}%
  \begin{tabular}{#2}#3\end{tabular}%
  \endgroup
}

\newenvironment{appendixtable}{\begin{table*}[!t]\centering}{\end{table*}}

\definecolor{cvprblue}{rgb}{0.21,0.49,0.74}
\usepackage[breaklinks,colorlinks,allcolors=cvprblue]{hyperref}
\hypersetup{
  pdftitle={Spectral Tail Auxiliary Learning for AI-Generated Image Detection},
  pdfauthor={Xingyi Li, Jiahui Zhang, Yiheng Li, Yun Cao, Wenhao Wang}
}

\title{Spectral Tail Auxiliary Learning for AI-Generated Image Detection}
\author{%
\begin{tabular}{c}
Xingyi Li\textsuperscript{1} \quad
Jiahui Zhang\textsuperscript{1} \quad
Yiheng Li\textsuperscript{2} \\
Yun Cao\textsuperscript{1}\thanks{Corresponding authors.} \quad
Wenhao Wang\textsuperscript{3}\footnotemark[1] \\
\small \textsuperscript{1}Institute of Information Engineering, Chinese Academy of Sciences, Beijing, China \\
\small \textsuperscript{2}Institute of Automation, Chinese Academy of Sciences, Beijing, China \\
\small \textsuperscript{3}Vast Intelligence Lab, Sydney, Australia \\
\small \texttt{\{lixingyi, zhangjiahui, caoyun\}@iie.ac.cn} \\
\small \texttt{liyiheng2024@ia.ac.cn} \quad
\small \texttt{wangwenhao@vastilab.com}
\end{tabular}
}

\begin{document}
\maketitle
\raggedbottom
\begin{abstract}
\label{abstract}
As generative image models evolved, the gap between generated and real images continues to narrow, making AI-generated image detection increasingly challenging. Many existing methods exploit frequency-domain cues for detection, typically described as frequency-domain artifacts or high-frequency discrepancies. However, the specific and recurring spectral regularities remain insufficiently understood and characterized. We systematically analyze the one-dimensional radial log-power spectra of real and generated images. We find that generated images do not necessarily exhibit higher or lower energy across the entire spectrum or high-band range. Instead, their spectra deviate from the power-law decay and show an anomalous uplift in the ultra-high-frequency tail. We term this phenomenon \emph{spectral tail uplift}. We further explain this phenomenon through nonlinear harmonic accumulation in trained generative models, suggesting that it can serve as a structural cue across generative architectures. Based on this observation, we propose Spectral Tail Auxiliary Learning (STAL), a frequency-domain auxiliary supervision framework for generalizable AI-generated image detection. STAL transfers spectral-tail cues from a tail-aware frequency teacher to a spatial detector during training, while all frequency-domain modules are discarded at inference time. Consequently, STAL introduces no inference overhead. Extensive experiments on 9 public datasets show that STAL achieves strong generalization and stability across generators, data distributions, and real-world scenarios.
\end{abstract}

\section{Introduction}
\label{Introduction}

\begin{figure*}[!t]
\centering
\includegraphics[width=0.9\linewidth]{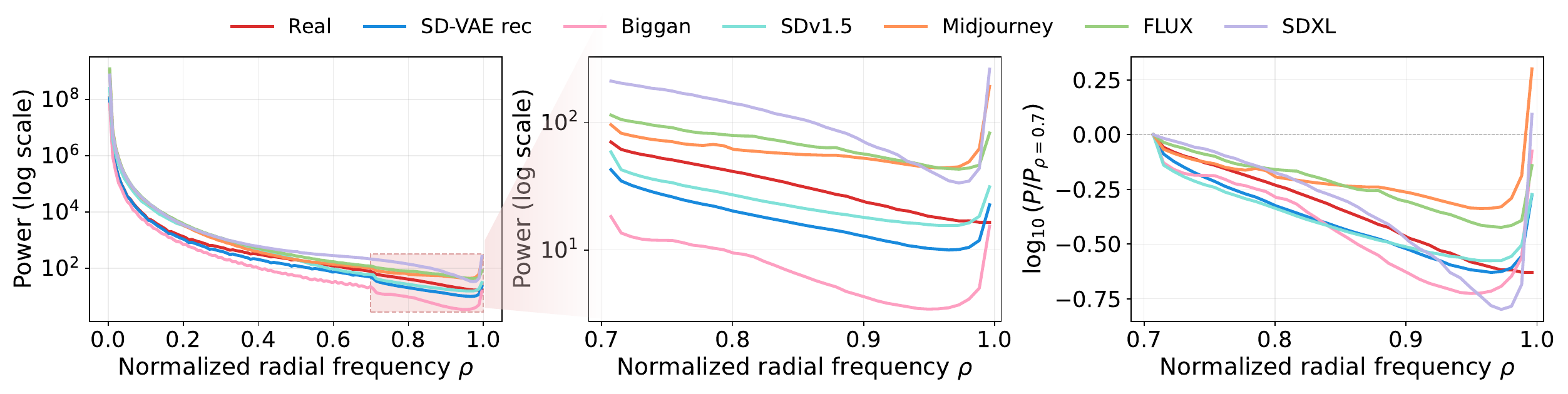}
\caption{Radial FFT \citep{Cooley1965FFT} power spectra of real and fake images. \textbf{Left:} spectra over the full radial frequency range. \textbf{Middle:} spectral tail over the local frequency range $\rho \in [0.7,1]$. \textbf{Right:} normalized tail curves anchored at $\rho=0.7$ to expose shape differences. Across generators, fakes show consistent spectral-tail deviations, revealing spectral tail uplift across architectures. The middle and right panels are included to help to understand and are not part of its definition.
}
\label{fig:multi_models_spec}
\end{figure*}

As generative image models continue to advance \citep{goodfellow2014GAN, Rombach2022LDM, bfl2024flux1dev, midjourney}, AI-generated images have become increasingly photorealistic, raising concerns about misuse and creating an need for detectors that are highly generalizable and remain reliable after the compression, resizing, and re-encoding that images undergo in the wild. Frequency-domain cues have long been regarded as a promising route toward both goals, since generative pipelines are believed to leave traces in the spectrum that are invisible to the eye \citep{Zhang2019, frank2020, Chu2025FIRE, Li2024Masksim}.


However, existing frequency-based detectors have several limitations. First, these spectral cues are loosely defined, often described as frequency-domain artifacts or high-frequency anomalies \citep{Zhang2019, Luo2024LARE2}, yet such cues do not exhibit a consistent pattern across generators. For example, Fig.~\ref{fig:multi_models_spec} shows that some generated images exhibit excessive high-frequency energy, while others show insufficient energy. Detectors that rely on such cues, therefore, struggle to achieve strong generalization across generators. Meanwhile, existing explanations for the origins of these artifacts are largely architecture-specific, focus mainly on GANs, and fail to provide a mechanism applicable to diffusion models and other modern generators. Moreover, most methods directly use spectral features at inference time, making them sensitive to perturbations. These limitations raise a more fundamental question: \textit{is there a spectral regularity that recurs across generators, can be explained by a common mechanism, and can be exploited without relying on fragile frequency-domain inputs at inference time?}
  
To answer this question, we systematically analyze the one-dimensional radial log-power spectra of real and generated images. We observe a phenomenon that consistently emerges across different generative architectures. As shown in Fig.~\ref{fig:multi_models_spec}, while real images follow an approximate power-law decay \citep{specdecayField, simoncelli2001natural} in the radial spectrum, generated images deviate from this behavior in the ultra-high-frequency tail, exhibiting an anomalous upward shift. We term this phenomenon \emph{spectral tail uplift}. Unlike spectral energy discrepancies, spectral tail uplift is a localized morphological cue that appears consistently across a wide range of GANs \citep{brock2018biggan}, diffusion models \citep{Rombach2022LDM, midjourney}, autoregressive models \citep{Han2025CVPR}, and reconstructions from different VAEs \citep{Rombach2022LDM}.

Controlled experiments further show that spectral tail uplift is associated with nonlinear activations in trained generative models. Removing nonlinear activations eliminates the uplift, while replacing trained weights with random initialization substantially weakens it. We explain this behavior from the perspective of harmonic accumulation. Pointwise nonlinear activations generate new harmonic components, while the learned parameters preserve and propagate them through the generative cascade. 
Because both components are shared by virtually all mainstream generative models, this mechanism provides an architecture-agnostic explanation for spectral tail uplift. This phenomenon is therefore a mechanistically grounded cross-generator characteristic rather than a generator- or dataset-specific artifact.

Based on this observation, we propose \textbf{S}pectral \textbf{T}ail \textbf{A}uxiliary \textbf{L}earning (\textbf{STAL}), a frequency-domain auxiliary supervision framework for generated-image detection. STAL leverages spectral tail uplift as a discriminative signal through training-time auxiliary supervision, enabling the detector to benefit from spectral-tail cues while maintaining robustness to post-processing operations at inference time. During training, STAL constructs a tail-aware frequency teacher and transfers spectral-tail cues to the representation of a spatial detector through auxiliary supervision. At inference time, all frequency-domain modules are discarded, and only the spatial detector is retained. 
In this way, STAL leverages tail uplift to improve detector generalization without introducing any additional inference cost. 
Our main contributions are summarized as follows:
\begin{itemize}[leftmargin=1.3em,itemsep=1pt,topsep=2pt]
\item We identify spectral tail uplift, a local spectral regularity in the ultra-high-frequency tail recurring across GANs, diffusion models, autoregressive models and VAEs.
\item We conduct a systematic analysis of spectral tail uplift and provide a cross-architecture explanation from the perspective of harmonic accumulation induced by nonlinear activations in trained generative models.
\item We propose Spectral Tail Auxiliary Learning, a auxiliary supervision framework based on spectral tail uplift, and validate its effectiveness on multiple public datasets, demonstrating significant improvements over state-of-the-art methods and greater robustness.
\end{itemize}

\section{Related Works}

\subsection{Generative Image Models}
Image synthesis technologies \citep{goodfellow2014GAN, ho2020DM, Rombach2022LDM, midjourney, SDXL} have advanced rapidly over the past few years. Early generative adversarial networks (GANs) \citep{goodfellow2014GAN, karras2019stylegan, zhu2017cyclegan, choi2018stargan} were already capable of producing photorealistic images, but they still suffered from issues such as mode collapse and training instability. Subsequently, \citet{ho2020DM} proposed Denoising Diffusion Probabilistic Models (DDPMs), which have since become the dominant paradigm in this field. Diffusion models generate high-quality images by simulating a thermodynamic diffusion process, starting from Gaussian noise and progressively recovering images through iterative denoising. \citet{Rombach2022LDM} further introduced Latent Diffusion Models (LDMs), which map the diffusion process into a low-dimensional latent space and employ a variational autoencoder (VAE) \citep{kingma2014VAE} as the encoder and decoder between pixel space and latent space, thereby substantially reducing computational costs while preserving generation quality. Many mainstream generative models \citep{SDXL, esser2024SD3, bfl2024flux1dev} developed thereafter have also adopted this paradigm, using VAEs for encoding and decoding.

\subsection{Frequency-based Generated Image Detection}

\begin{figure*}[!t]
\centering
\includegraphics[width=0.9\linewidth]{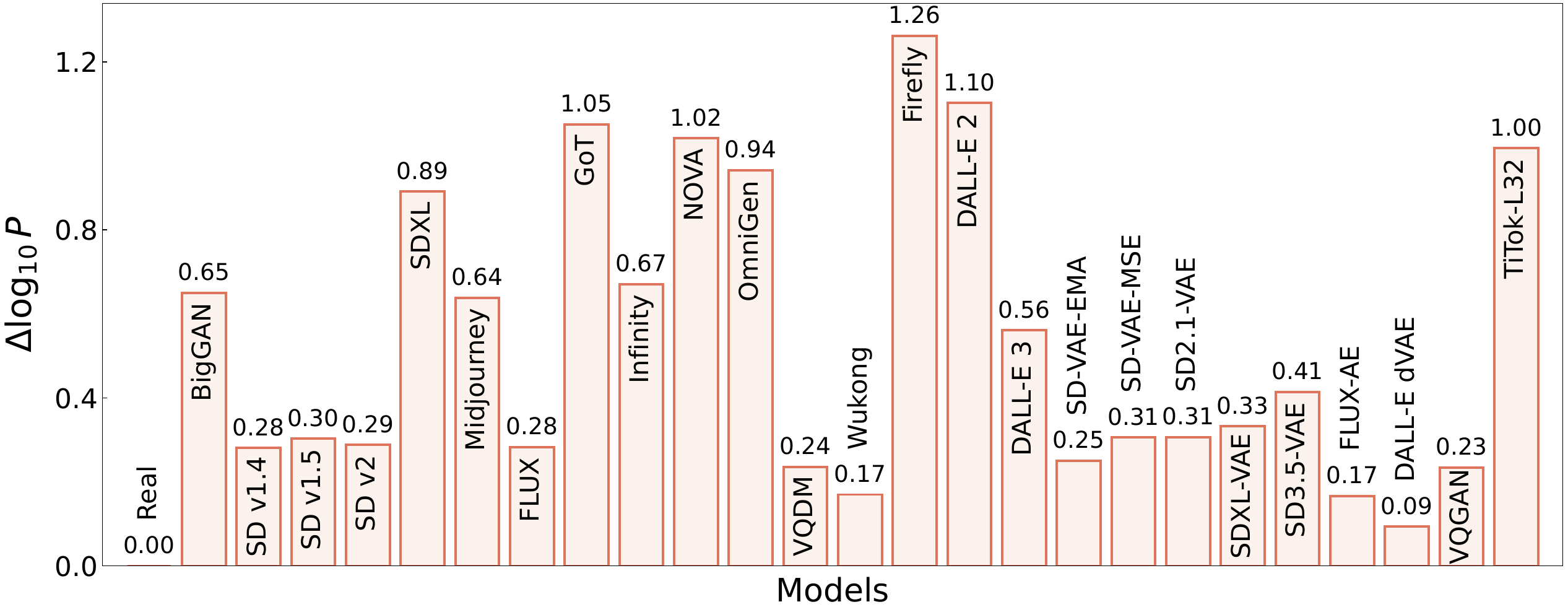}
\caption{Spectral tail uplift across diverse generative architectures. Real images exhibit essentially no uplift, whereas images from a wide range of generators consistently show positive uplift. }
\label{fig:delta_log10_p_bar}
\end{figure*}

Many studies \citep{frank2020, Tan2024FreqNet, Liu2024fatformer, Luo2024LARE2, yan2025AIDE} have explored frequency-domain cues for AI-generated image detection. \citet{Zhang2019} showed that up-sampling operations in GANs introduce periodic frequency-domain artifacts. \citet{Chandrasegaran2021CVPR} investigated high-frequency spectral decay discrepancies in GAN-generated images and showed that they can be largely eliminated by replacing zero-insertion-based scaling in the generator's final upsampling stage, thereby yielding spectral decay that is more consistent with that of real images. However, we study a localized non-monotonic upturn at the extreme end of the spectrum, associated with the accumulation of harmonics induced by nonlinear activations in trained generators and still observable across a variety of generative models whose final upsampling stage does not employ zero-insertion-based scaling. Some works exploit frequency information directly or jointly with spatial features. FreqNet \citep{Tan2024FreqNet} learns in frequency space to improve generalization, while MaskSim \citep{Li2024Masksim} learns discriminative spectral masks. \citet{Luo2024LARE2} leverages high-frequency reconstruction errors, and FIRE \citep{Chu2025FIRE} focuses on mid-frequency components that generators struggle to reconstruct. Other methods combine frequency cues with pretrained CLIP \citep{Radford2021CLIP} features \citep{yan2025AIDE, Liu2024fatformer}. Although prior works leverage frequency cues in various ways, they mostly rely on generator-dependent spectral discrepancies, leaving a recurring cross-generator spectral regularity with a common mechanism insufficiently characterized and exploited.

\section{Method}
\subsection{Spectral Tail Uplift}

We compare the azimuthally averaged $1$D radial power spectra \(S(\rho)\) of real and generated images, where \(\rho \in [0,1]\) denotes the normalized radial frequency. As shown in Fig.~\ref{fig:multi_models_spec}, real-image spectra approximately follow a power-law decay,
\begin{equation}
    S(\rho) \propto \rho^{-\alpha},
    \label{eq:power_law_decay}
\end{equation}
and decrease monotonically up to \(\rho=1\). Generated images exhibit similar behavior at low and mid frequencies, but show a characteristic decrease-to-increase transition in the ultra-high-frequency tail. We refer to this pattern as \emph{spectral tail uplift}, which definition is as follow: given $\rho_0$, $S$ exhibits spectral tail uplift if there exists $\rho^{}\in[\rho_0,1)$ such that $S(\rho)$ decreases on $[\rho_0,\rho^{}]$ and increases on $[\rho^{*},1]$. We quantify its strength by
\begin{equation}
\Delta
=
\log_{10}S(1)
-
\min_{\rho\geq\rho_0}\log_{10}S(\rho),
\end{equation}
where $\Delta=0$ for a monotonically decreasing spectrum and $\Delta>0$ when an uplift is present. We use $\Delta$ only for quantitative comparisons in Figs.~\ref{fig:delta_log10_p_bar}, Figs.~\ref{fig:VAE_activation} and Appendix~\ref{app:Controlled_Experiments}. 
Fig.~\ref{fig:delta_log10_p_bar} shows the pattern across GANs \citep{brock2018biggan}, diffusion models \citep{Rombach2022LDM, midjourney}, autoregressive models \citep{Han2025CVPR}, and reconstructions from different VAEs \citep{Rombach2022LDM}, but not in real images.

\subsection{Analysis}
\label{theory analysis}

\begin{figure*}[!t]
\centering
\includegraphics[width=0.9\linewidth]{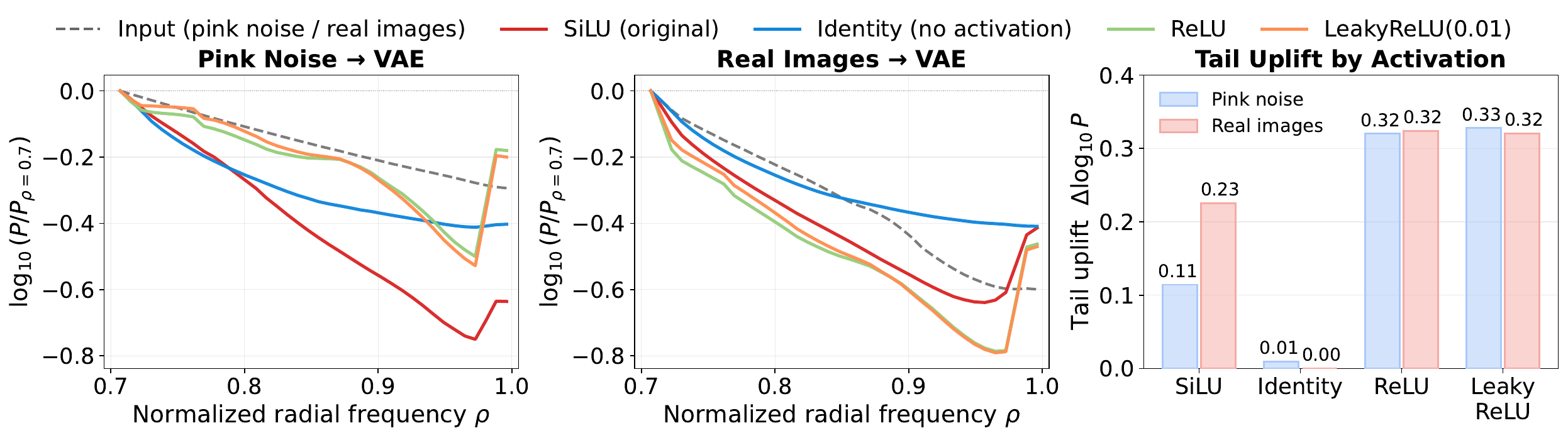}
\caption{Activation nonlinearity is necessary for the spectral tail uplift. We replace SiLU in SD-VAE with Identity, ReLU, or LeakyReLU, then pass pink noise (\textbf{left}) and real images (\textbf{middle}) through the modified VAE. Normalized Curves show spectra on $\rho\in[0.7,1]$. \textbf{Right:} tail uplift $\Delta\log_{10}P$, the rise from the tail's minimum to $\rho=1$. Identity collapses the uplift to $\approx 0$. ReLU and LeakyReLU strengthen it.}
\label{fig:VAE_activation}
\end{figure*}

The observation above raises a natural question: where does the extra energy at the extreme high-frequency tail come from? We abstract a typical generative decoder as an $L$-layer cascade, in which each layer applies a trained convolution followed by a pointwise nonlinear activation,
\begin{equation}
x_\ell \;=\; \phi\!\left( H_\ell \ast x_{\ell-1} \right), \qquad \ell = 1, \dots, L,
\label{eq:cascade}
\end{equation}
where $H_\ell$ denotes the convolution at layer~$\ell$, and $\phi$ is a pointwise nonlinear activation. We omit attention, normalization, residual connections, and upsampling from this abstraction, which isolates the single mechanism analyzed below. Within this cascade, the linear convolution can only rescale the amplitude of frequencies already present in its input, but cannot create new frequencies. The pointwise nonlinear activation, in contrast, introduces frequency components that were absent from its input. We now analyze how the emergence of spectral tail uplift is closely related to high-frequency harmonics generated by nonlinear activations and propagated through the learned convolutions.

\noindent\textbf{Harmonic generation.}
Consider a bandlimited input
\begin{equation}
x(t) \;=\; \sum_{m=-M}^{M} \hat{x}_m \, e^{i m t},
\label{eq:input}
\end{equation}
of bandwidth $M$, with Fourier coefficients $\hat{x}_m$ and a nonzero top component $\hat{x}_M \neq 0$. Let $\phi(z) = \sum_{q=0}^{d} a_q z^q$ be a degree-$d$ polynomial activation with $a_d \neq 0$ and $d \geq 2$. 

\noindent\textbf{Theorem 1 (Harmonic generation). }
The highest positive frequency of $\phi(x)$ is extended from $M$ to $dM$, and the Fourier coefficient at frequency $dM$ is exactly
\begin{equation}
\widehat{\phi(x)}_{dM} \;=\; a_d \, (\hat{x}_M)^{d}.
\label{eq:topcoef}
\end{equation}
This follows from the fact that $x^d$ corresponds to a $d$-fold self-convolution in the Fourier domain. The full proof is given in Appendix~\ref{app:harmonic_generation}. In other words, each nonlinear activation layer introduces new frequency components beyond those present in the input, thereby extending the signal toward higher frequencies.

Natural images may also undergo nonlinear operations inside the camera image signal processing (ISP) pipeline, including gamma correction and denoising. The key difference is that these operations are constrained by optical imaging, sensor sampling, and camera processing and are not organized as a learned convolution-activation cascade. As a result, the spectra of real images retain an approximate power-law decay. A generative decoder, by contrast, repeatedly interleaves trained convolutions with pointwise nonlinear activations. Each pointwise activation can generate new harmonic components, and the learned convolutional filters carry these harmonics toward progressively higher frequencies. This repeated synthesis provides the conditions for the sustained accumulation of high-frequency harmonics, which provides a concrete route by which extra energy reaches the spectral tail of generated images.

\begin{figure*}[t]
\centering
\includegraphics[width=0.9\linewidth]{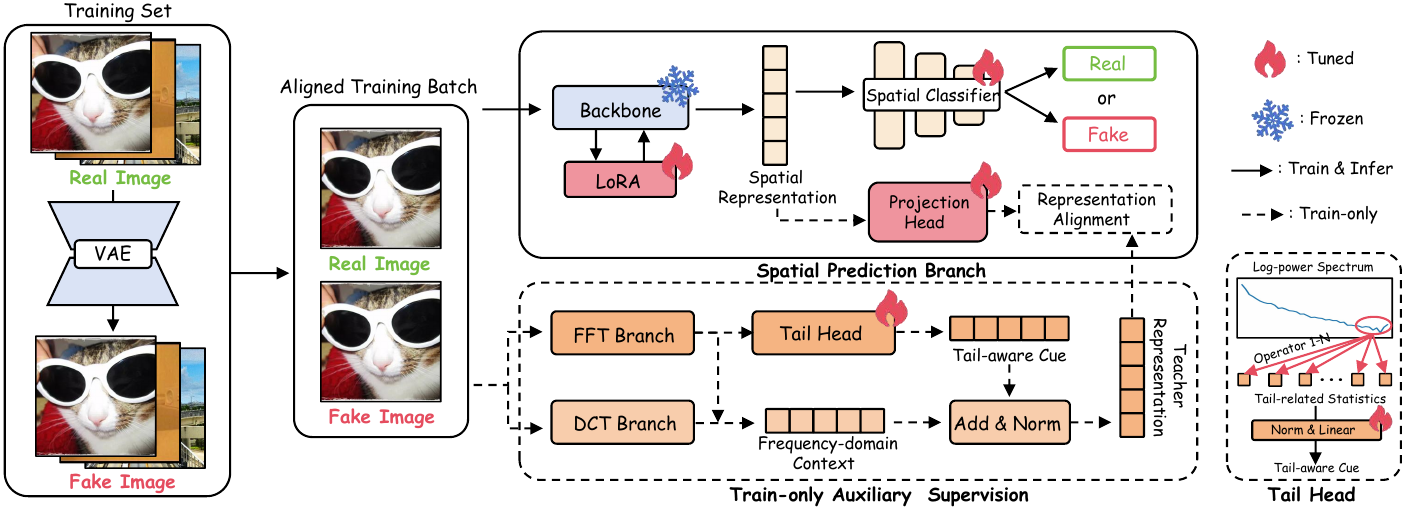}
\caption{
Overview of STAL. A tail-aware frequency teacher extracts spectral-tail cues from a frequency-preserving view and aligns them with projected spatial features. At inference, the frequency modules and projection head are discarded, leaving only the spatial detector.
}
\label{fig:model}
\end{figure*}

\noindent\textbf{Harmonic propagation through the cascade.}
In the polynomial model above, each nonlinearity extends the accessible top frequency by a factor of $d$. Let the initial frequency be $k_0$, there exists a highest-order harmonic path that evolves as $k_0 \to d k_0 \to d^{2} k_0 \to \cdots \to d^{L} k_0$. 

\noindent\textbf{Theorem 2 (Harmonic-chain propagation). }
For a single-tone input $x_0(t) = A\cos(k_0 t)$ of amplitude $A$ and frequency $k_0$, if the filter responses along this path are nonzero, i.e., $H_\ell(d^{\ell-1}k_0)\neq 0$ for all $\ell=1,\dots,L$, the power at the top harmonic $d^{L} k_0$ at layer $L$ is
\begingroup
\footnotesize
\begin{equation}
\bigl| \hat{x}_L(d^L k_0) \bigr|^{2}
\;=\;
\lvert a_d \rvert^{\,\frac{2(d^{L}-1)}{d-1}}
\left(\frac{A}{2}\right)^{\!2 d^{L}}
\prod_{\ell=1}^{L}
\bigl\lvert H_\ell\!\left(d^{\ell-1} k_0\right) \bigr\rvert^{\,2 d^{L-\ell+1}}.
\label{eq:toppower}
\end{equation}
\endgroup
The proof is given in Appendix~\ref{app:harmonic_chain}. The top-harmonic power thus depends jointly on the nonlinearity strength $\lvert a_d \rvert$, the input amplitude $A$, and the filter gains $\lvert H_\ell \rvert$ evaluated along the harmonic chain. Taking logarithms exposes the per-layer contributions more clearly:
\begin{equation}
\log \bigl| \hat{x}_L(d^L k_0) \bigr|^{2}
\;=\; C_0 \;+\; \sum_{\ell=1}^{L} 2\, d^{L-\ell+1} \, \log \bigl\lvert H_\ell\!\left(d^{\ell-1} k_0\right) \bigr\rvert,
\label{eq:logpower}
\end{equation}
where $C_0$ absorbs the contributions of $\lvert a_d \rvert$ and $A$. The weights $2 d^{L-\ell+1}$ scale exponentially with the remaining depth, so the top-harmonic path is highly sensitive to the filter gain of every layer along the chain. The nonlinearity injects new high-frequency content at each layer, and the cascaded convolutions propagate this content along the harmonic chain toward progressively higher frequencies, where it emerges as the dominant contribution at the high-frequency tail and produces the uplift we observe.

Our controlled experiments support the above analysis: tail uplift consistently appears in trained generative models, while it is almost completely eliminated by removing nonlinear activations as shown in Fig.~\ref{fig:VAE_activation}, and substantially suppressed when the model uses untrained random weights. Complete experiment results are provided in Appendix~\ref{app:Controlled_Experiments}. Since nearly all mainstream generative models, including diffusion models, GANs, and autoregressive models with visual decoders, are trained and incorporate nonlinear activations, we argue that tail uplift recurs across generators, as we verify on GAN,
diffusion, autoregressive models, as well as reconstructions from different VAE/AE architectures, and serves as a generator-agnostic cue.

\subsection{Spectral Tail Auxiliary Learning}
\label{sec:model}

We propose Spectral Tail Auxiliary Learning (STAL), as shown in Fig.~\ref{fig:model}, which transfers spectral-tail cues to a spatial detector through a training-only frequency teacher. The spatial view $x_s$, constructed following DDA \citep{chen2026DDA}, is processed by the visual encoder $E_{\theta}$ and classifier $C_s$. In parallel, a frequency-preserving view $x_f$ shares the same geometric transformations as $x_s$ but excludes augmentations that distort spectral structure. Its radial log-power spectra $S_{\mathrm{rad}}(x_f)$ are encoded by a tail head $T_{\omega}$ and a context encoder $F_{\psi}$, capturing spectral-tail uplift and global spectral context, respectively. Their representations form the teacher target $t_f$, which is aligned with the projected spatial representation during training. At inference, all frequency-domain modules and the projection head are removed, leaving only the spatial detector.

\noindent\textbf{Tail-Aware Frequency Teacher.}
\textbf{(i)Tail Head.}
We design a tail head $T_{\omega}$ to characterize the pattern of spectral tail uplift. The input to $T_{\omega}$ is restricted to the segment of the radial spectrum $S_{\mathrm{rad}}(x_f)$ over the normalized radial frequency range $\rho \in [0.7, 1.0]$. A predefined operator set
$\mathcal{P} = \{p_1, p_2, ..., p_n\}$
is employed to extract a collection of tail-related characteristics, capturing the spectral morphology from the turning point to subsequent variations in spectral slope. The resulting representation is further processed by layer normalization and a linear projection to obtain the tail embedding $e_t$, while a linear classifier produces an auxiliary prediction $\hat{y}_t$:
\begin{equation}
    e_t,\, \hat{y}_t
    =
    T_{\omega}\!\left(
        \left.
        S_{\mathrm{rad}}(x_f)
        \right|_{\rho \in [0.7,\,1.0]}
    \right).
\end{equation}
\textbf{(ii)Full-Spectrum Context.}
Since spectral tail uplift is a localized spectral pattern, its abnormality is more effectively characterized with reference to the overall trend across the remaining frequency bands. We therefore employ a lightweight frequency-domain encoder $F_{\psi}$ to encode the complete spectrum $S_{\mathrm{rad}}(x_f)$ together with the local DCT statistics $D_{\mathrm{loc}}(x_f)$ into a contextual representation:
\begin{equation}
    h_f
    =
    F_{\psi}\!\left(
        S_{\mathrm{rad}}(x_f),
        D_{\mathrm{loc}}(x_f)
    \right).
\end{equation}
Here, $D_{\mathrm{loc}}(x_f)$ preserves local frequency-distribution information that may be suppressed by radial averaging, thereby complementing the global spectral trend represented by $S_{\mathrm{rad}}(x_f)$. \textbf{(iii)Teacher Target.}
We combine the contextual representation $h_f$ with the tail embedding $e_t$ to construct the frequency-domain supervision target:
\begin{equation}
    t_f
    =
    \operatorname{LN}\!\left(
        h_f + \beta e_t
    \right),
\end{equation}
where $\operatorname{LN}(\cdot)$ denotes layer normalization, and $\beta$ controls the contribution of the tail embedding. The two components of the teacher thus play complementary roles. The tail head captures the localized decay profile of the spectral tail, whereas the frequency-domain encoder provides the global spectral trend against which this local pattern is interpreted. Consequently, $t_f$ represents the shape of the spectral tail relative to the global spectral trend.

\noindent\textbf{Frequency-to-spatial auxiliary learning.}
The frequency teacher is designed to transfer tail-aware frequency cues to the spatial representation through training-time auxiliary supervision. The main spatial branch extracts an image representation from the spatial view $x_s$ and produces the classification prediction:
\begin{equation}
    \mathbf{v}_s = E_\theta(x_s), \qquad
    \hat{y}_s = C_s(\mathbf{v}_s),
\end{equation}
where $E_\theta$ denotes the visual encoder of the spatial detector, $\mathbf{v}_s$ is the spatial image representation, and $C_s$ is the spatial binary classifier. To receive supervision from the frequency teacher, we use a projection head $g_\eta$ to map the spatial representation into the teacher feature space:
\begin{equation}
    \mathbf{p}_s = g_\eta(\mathbf{v}_s).
\end{equation}
The projected spatial feature $\mathbf{p}_s$ is used only for training-time representation alignment and is discarded at inference time. We then align $\mathbf{p}_s$ with the stop-gradient frequency teacher target:
\begin{equation}
    \mathcal{L}_{\mathrm{align}}
    =
    \frac{1}{2}
    \sum_{c\in\{0,1\}}
    \mathbb{E}_{i:y_i=c}
    \left[
    1 -
    \cos(\mathbf{p}_{s,i}, \mathrm{sg}(\mathbf{t}_{f,i}))
    \right].
\end{equation}
Here, $y_i\in\{0,1\}$ denotes the authenticity label of the $i$-th sample, $\mathbb{E}_{i:y_i=c}$ denotes averaging over samples of class $c$ in the current mini-batch, and $\mathrm{sg}(\cdot)$ denotes stop-gradient operation. This loss computes the alignment error separately for real and generated images and assigns equal weight to the two classes, preventing the alignment from being dominated by one class. The stop-gradient operation fixes the teacher-side target and prevents the alignment loss from modifying the frequency supervisory signal. By imposing this representation constraint only during training, STAL transfers spectral-tail cues to the spatial detector without requiring any frequency-domain input or teacher branch at inference time. In addition to representation alignment, we apply auxiliary classification losses to the frequency teacher to ensure that it remains discriminative:
\begin{equation}
    \hat{y}_f = C_f(\mathbf{h}_f), \
    \mathcal{L}_{\mathrm{freq}} =
    \mathrm{BCE}(\hat{y}_f, y),
    \
    \mathcal{L}_{\mathrm{tail}} =
    \mathrm{BCE}(\hat{y}_t, y).
\end{equation}
Here, $C_f$ is an auxiliary classifier applied to the frequency context representation, and $\hat{y}_f$ is the prediction from the frequency context branch. $\mathcal{L}_{\mathrm{freq}}$ encourages $\mathbf{h}_f$ to preserve task-relevant frequency context, while $\mathcal{L}_{\mathrm{tail}}$ encourages the tail head to learn spectral-tail structures that are useful for generated-image detection. The frequency auxiliary objective is defined as
\begin{equation}
    \mathcal{L}_{\mathrm{aux}}
    =
    \lambda_{\mathrm{freq}} \mathcal{L}_{\mathrm{freq}}
    +
    \lambda_{\mathrm{align}} \mathcal{L}_{\mathrm{align}}
    +
    \lambda_{\mathrm{tail}} \mathcal{L}_{\mathrm{tail}},
\end{equation}
where $\lambda_{\mathrm{freq}}$, $\lambda_{\mathrm{align}}$, and $\lambda_{\mathrm{tail}}$ are the weights of corresponding loss weights. The spatial branch is optimized with a classification loss and a supervised contrastive loss, following the loss configuration as DDA \citep{chen2026DDA}:
\begin{equation}
    \mathcal{L}_{\mathrm{cls}} =
    \mathrm{BCE}(\hat{y}_s, y),
    \qquad
    \mathcal{L}_{\mathrm{spatial}}
    =
    \lambda_{\mathrm{cls}} \mathcal{L}_{\mathrm{cls}}
    +
    \lambda_{\mathrm{con}} \mathcal{L}_{\mathrm{con}},
\end{equation}
where $\mathcal{L}_{\mathrm{con}}$ denotes the supervised contrastive loss, and $\lambda_{\mathrm{cls}}$ and $\lambda_{\mathrm{con}}$ are the corresponding loss weights. The overall training objective is
\begin{equation}
    \mathcal{L}
    =
    \mathcal{L}_{\mathrm{spatial}}
    +
    w_{\mathrm{aux}}(\tau)\mathcal{L}_{\mathrm{aux}},
\end{equation}
where $w_{\mathrm{aux}}(\tau)$ is a curriculum weight that controls the strength of the frequency auxiliary terms according to the training progress $\tau$. The frequency auxiliary supervision primarily shapes the spatial representation in the early and middle stages of training. Subsequently, as training proceeds, its weight gradually decreases, reducing the risk that the final detector becomes overly dependent on the training-time frequency teacher.

\noindent\textbf{Inference.}
At inference time, we discard all frequency-domain modules, including the frequency branch, the tail head, and the projection head used for representation alignment. And then, the final prediction is produced solely by the spatial detector:
\begin{equation}
    \hat{y} = C_s(E_\theta(x)).
\end{equation}
Thus, STAL uses spectral tail uplift only as a training-time auxiliary supervisory signal, introducing no frequency-domain input, teacher branch, or additional computational cost at inference time.

\section{Experiments}
\label{Experiments}
\begin{table*}[t!]
    \centering
    \caption{Overall comparison across 9 benchmarks. Balanced accuracy (\%) is reported. DDA* denotes DDA \cite{chen2026DDA} with DINOv3-H+ \cite{Simeoni2025DINOv3} backbone. The best and second-best results are highlighted in \textbf{bold} and \underline{underline}. Numbers below each dataset name indicate the number of generators, where G, D, and AR denote GANs, diffusion models, and auto-regressive models, respectively.}
    \vspace{1mm}
    \small
    \setlength{\tabcolsep}{2pt}
    \renewcommand{\arraystretch}{0.98}
    \resizebox{\textwidth}{!}{%
    \begin{tabular}{@{}lcccccccccc@{}}
    \toprule
    \multirow{3}{*}{\textbf{Method}}
      & \multicolumn{6}{c}{\textbf{Standard Benchmarks}}
      & \multicolumn{3}{c}{\textbf{In-the-Wild Datasets}}
      & \multirow{3}{*}{\textbf{Avg}} \\
    \cmidrule(lr){2-7} \cmidrule(lr){8-10}
      & \textbf{GenImage}
      & \textbf{DRCT-2M}
      & \textbf{EvalGEN}
      & \textbf{Synthbuster}
      & \textbf{ForenSynths}
      & \textbf{\makecell{AIGCDetection\\Benchmark}}
      & \textbf{Chameleon}
      & \textbf{SynthWildx}
      & \textbf{WildRF}
      & \\
      & 1G + 7D
      & 13D
      & 3D + 2AR
      & 9D
      & 13G
      & 7G + 10D
      & Unknown
      & 3D
      & Unknown
      & \\
    \midrule

    NPR (CVPR'24)~\cite{Tan2024NPR}
      & 51.5 $\pm$ 6.3
      & 30.4 $\pm$ 2.7
      & 2.9 $\pm$ 2.7
      & 50.0 $\pm$ 2.6
      & 47.9 $\pm$ 22.6
      & 53.1 $\pm$ 12.2
      & 59.9
      & 49.8 $\pm$ 10.0
      & 63.5 $\pm$ 13.6
      & 45.4 $\pm$ 18.4 \\

    \addlinespace[1pt]
    UnivFD (CVPR'23)~\cite{Ojha2023UnivFD}
      & 64.1 $\pm$ 10.8
      & 62.6 $\pm$ 9.5
      & 15.4 $\pm$ 14.2
      & 67.8 $\pm$ 14.4
      & 77.7 $\pm$ 16.1
      & 72.5 $\pm$ 17.3
      & 50.7
      & 52.3 $\pm$ 11.3
      & 55.3 $\pm$ 5.7
      & 57.6 $\pm$ 18.2 \\

    \addlinespace[1pt]
    FatFormer (CVPR'24)~\cite{Liu2024fatformer}
      & 62.8 $\pm$ 10.4
      & 50.1 $\pm$ 3.6
      & 45.6 $\pm$ 33.1
      & 56.1 $\pm$ 10.7
      & 90.0 $\pm$ 11.8
      & 85.0 $\pm$ 14.9
      & 51.2
      & 52.1 $\pm$ 8.2
      & 58.9 $\pm$ 8.0
      & 61.3 $\pm$ 15.7 \\

    \addlinespace[1pt]
    SAFE (KDD'25)~\cite{Li2025SAFE}
      & 50.3 $\pm$ 1.2
      & 50.4 $\pm$ 1.3
      & 1.1 $\pm$ 0.6
      & 46.5 $\pm$ 20.8
      & 49.7 $\pm$ 2.7
      & 50.3 $\pm$ 1.1
      & 59.2
      & 49.1 $\pm$ 0.7
      & 57.2 $\pm$ 18.5
      & 46.0 $\pm$ 17.3 \\

    \addlinespace[1pt]
    C2P-CLIP (AAAI'25)~\cite{Tan2025C2PCLIP}
      & 74.4 $\pm$ 8.4
      & 58.9 $\pm$ 10.8
      & 38.9 $\pm$ 31.2
      & 68.5 $\pm$ 11.4
      & 92.0 $\pm$ 10.1
      & 81.4 $\pm$ 15.6
      & 51.1
      & 57.1 $\pm$ 4.2
      & 59.6 $\pm$ 7.7
      & 64.7 $\pm$ 16.2 \\

    \addlinespace[1pt]
    AIDE (ICLR'25)~\cite{yan2025AIDE}
      & 61.2 $\pm$ 11.9
      & 61.2 $\pm$ 8.6
      & 19.1 $\pm$ 11.1
      & 53.9 $\pm$ 18.6
      & 59.4 $\pm$ 24.6
      & 63.6 $\pm$ 13.9
      & 63.1
      & 48.8 $\pm$ 0.8
      & 58.4 $\pm$ 12.9
      & 54.3 $\pm$ 14.0 \\

    \addlinespace[1pt]
    DRCT (ICML'24)~\cite{Chen2024DRCT}
      & 84.7 $\pm$ 2.7
      & 94.4 $\pm$ 1.8
      & 77.8 $\pm$ 5.4
      & 84.8 $\pm$ 3.6
      & 73.9 $\pm$ 13.4
      & 81.4 $\pm$ 12.2
      & 56.6
      & 55.1 $\pm$ 1.8
      & 50.6 $\pm$ 3.5
      & 73.3 $\pm$ 15.5 \\

    \addlinespace[1pt]
    AlignedForensics (ICLR'25)~\cite{rajan2025Aligned}
      & 79.0 $\pm$ 22.7
      & 95.1 $\pm$ 6.5
      & 68.0 $\pm$ 20.7
      & 77.4 $\pm$ 25.0
      & 53.9 $\pm$ 7.1
      & 66.6 $\pm$ 21.6
      & 71.0
      & 78.8 $\pm$ 17.8
      & 80.1 $\pm$ 10.3
      & 74.4 $\pm$ 11.4 \\

    \addlinespace[1pt]
    DDA (NeurIPS'25)~\cite{chen2026DDA}
      & 91.7 $\pm$ 7.8
      & 98.0 $\pm$ 1.4
      & 97.2 $\pm$ 4.2
      & 90.1 $\pm$ 5.6
      & 81.4 $\pm$ 13.9
      & \underline{87.8 $\pm$ 12.6}
      & 82.4
      & 90.9 $\pm$ 3.1
      & 90.3 $\pm$ 3.5
      & 90.0 $\pm$ 5.7 \\

    \midrule
    DDA*
      & \underline{97.4 $\pm$ 0.4}
      & \underline{99.7 $\pm$ 0.1}
      & \textbf{99.3 $\pm$ 0.8}
      & \textbf{99.8 $\pm$ 0.1}
      & 83.8 $\pm$ 6.7
      & \underline{92.4 $\pm$ 7.0}
      & \underline{92.1}
      & \underline{92.0 $\pm$ 0.3}
      & \underline{96.3 $\pm$ 0.7}
      & \underline{94.2 $\pm$ 5.8} \\

    \addlinespace[1pt]
    \rowcolor[HTML]{E3F1FF}
    \textbf{STAL (ours)}
      & \textbf{98.6 $\pm$ 0.9}
      & \textbf{99.8 $\pm$ 0.1}
      & \underline{98.9 $\pm$ 1.8}
      & \underline{99.7 $\pm$ 0.4}
      & \textbf{94.7 $\pm$ 5.0}
      & \textbf{96.9 $\pm$ 4.2}
      & \textbf{92.2}
      & \textbf{95.2 $\pm$ 0.4}
      & \textbf{97.9 $\pm$ 1.0}
      & \textbf{97.0 $\pm$ 2.6} \\
    \bottomrule
    \end{tabular}%
    }
    \label{tab:allbench}
    \vspace{-1mm}
  \end{table*}

\subsection{Experimental Setup}

\noindent\textbf{Datasets.} We evaluate STAL on nine public benchmarks, including six standard benchmarks (GenImage \citep{Zhu2023GenImage}, DRCT-2M \citep{Chen2024DRCT}, Synthbuster \citep{Synthbuster}, EvalGEN \citep{chen2026DDA}, AIGCDetectionBenchmark \citep{Zhong2023PatchCraft}, and ForenSynths \citep{Wang2020CNNspot}) and three in-the-wild datasets (Chameleon \citep{yan2025AIDE}, SynthWildx \citep{SynthWildx}, and WildRF \citep{WildRF}). These datasets cover diverse real image sources and synthetic images produced by GANs, diffusion models, and autoregressive models, while the in-the-wild datasets are collected from open-world scenarios, which involve unknown generators and post-processing operations.

\noindent\textbf{Evaluation Metrics and Comparative Methods.}
We report balanced accuracy (BAL), defined as the average of real image and generated image accuracies. For benchmarks with multiple subsets, we report the arithmetic mean over subsets. We compare STAL with existing representative methods, including NPR \citep{Tan2024NPR}, UnivFD \citep{Ojha2023UnivFD}, FatFormer \citep{Liu2024fatformer}, C2P-CLIP \citep{Tan2025C2PCLIP}, SAFE \citep{Li2025SAFE}, AIDE \citep{yan2025AIDE}, DRCT \citep{Chen2024DRCT}, AlignedForensics \citep{rajan2025Aligned}, and DDA \citep{chen2026DDA}, using their publicly released model weights. We also include DDA*, a DDA variant trained under the same setting as STAL with DINOv3-H+ \citep{Simeoni2025DINOv3} as the backbone.

\noindent\textbf{Implementation Details.} During training, we use real images from MSCOCO \citep{MSCOCO} and their reconstructed images as the training data. For the spatial input, we follow DDA \citep{chen2026DDA} to construct aligned training samples. In parallel, we keep the original input view for the frequency auxiliary branch to avoid disrupting frequency-domain characteristics. We train the model for one epoch with a batch size of 16 per GPU. The spatial detector uses DINOv3-H+ \citep{Simeoni2025DINOv3} as the backbone and adopts LoRA \citep{hu2022lora} with a rank of 8 and $\alpha=1.0$ for parameter-efficient fine-tuning. The input resolution is $336 \times 336$. We use AdamW \citep{Adamw} with a learning rate of $1\times10^{-4}$ and a weight decay of 0.005. More implementation details are provided in Appendix~\ref{app:implementation}.

\subsection{Cross-Dataset Comparison}
\label{Cross-Dataset Comparison}

Table~\ref{tab:allbench} presents the results on nine benchmarks. Compared with existing methods, STAL achieves the best average BAL of 97.0\% , ranks first on 7 out of 9 datasets, and shows stronger stability across datasets. Compared with DDA*, STAL improves the average BAL by 2.8 \% and reduces the cross-dataset standard deviation from 5.8 to 2.6. The gains are especially clear on ForenSynths, AIGCDetectionBenchmark, and SynthWildx, suggesting that the improvement is not merely due to a stronger backbone.

\subsection{Robustness Analysis}
\begin{figure*}[t]
\centering
\includegraphics[width=0.9\linewidth]{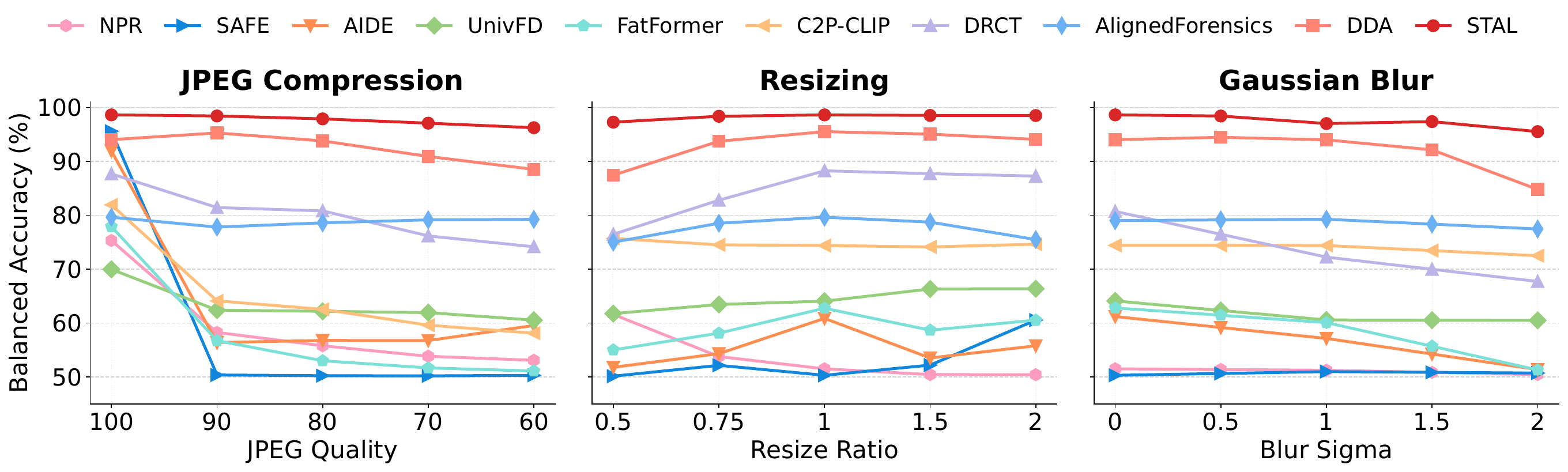}
\caption{Robustness analysis on GenImage. We evaluate STAL and competing methods under JPEG compression, resizing, and Gaussian blur with increasing perturbation strengths.}
\label{fig:robustness}
\end{figure*}

Fig.~\ref{fig:robustness} evaluates robustness on GenImage under JPEG compression, resizing, and Gaussian blur. STAL ranks first across all perturbation settings. Its balanced accuracy drops by only 2.38 points when JPEG quality decreases from $Q=100$ to $Q=60$, by 1.35 points under strong downsampling ($\alpha=0.5$), and by at most 3.10 points under Gaussian blur. The margins over the second-best method remain large under severe perturbations, reaching 7.73 points at $Q=60$, 9.84 points at $\alpha=0.5$, and 10.73 points at $\sigma=2.0$.

\subsection{Ablation Analysis}
\begin{table}[t]
\centering
\caption{Frequency usage ablation on 9 benchmarks. The rows compare the spatial-only baseline, direct spatial-frequency inference, and training-time auxiliary frequency supervision (STAL), respectively.  Average balanced accuracy (\%) is reported, and $\Delta$ is relative to the spatial-only baseline.}
\label{tab:ablation}

\small
\setlength{\tabcolsep}{8pt}
\renewcommand{\arraystretch}{1.05}

\resizebox{0.9\linewidth}{!}{%
  \begin{tabular}{ccccc}
    \toprule
    \textbf{\makecell{Spatial\\branch}} &
    \textbf{\makecell{Frequency\\inference}} &
    \textbf{\makecell{Auxiliary\\supervision}} &
    \textbf{BAL (\%)} &
    \textbf{$\Delta$} \\
    \midrule
    \cmark & \xmark & \xmark
      & 94.2 & -- \\
    \cmark & \cmark & \xmark
      & 96.0 & +1.8 \\
    \rowcolor[HTML]{E3F1FF}
    \cmark & \xmark & \cmark
      & \textbf{97.0} & \textbf{+2.8} \\
    \bottomrule
  \end{tabular}%
}
\end{table}
\begin{table}[t]
    \centering
    \caption{Frequency branch leave-one-out ablation on 9 benchmarks. Check marks indicate the teacher components used for auxiliary supervision. The first row is the spatial-only baseline. The highlighted row is STAL.}
    \label{tab:frequency_components}

    \small
    \setlength{\tabcolsep}{10pt}
    \renewcommand{\arraystretch}{1.05}

    \resizebox{0.9\linewidth}{!}{%
      \begin{tabular}{cccc}
        \toprule
        \textbf{Frequency context} &
        \textbf{Tail head} &
        \textbf{BAL (\%)} &
        \textbf{$\Delta$} \\
        \midrule
        \xmark & \xmark & 94.2 & -- \\
        \cmark & \xmark & 94.4 & +0.2 \\
        \xmark & \cmark & 96.1 & +1.9 \\
        \rowcolor[HTML]{E3F1FF}
        \cmark & \cmark & \textbf{97.0} & \textbf{+2.8} \\
        \bottomrule
      \end{tabular}%
    }
\end{table}
  \begin{table}[t]
    \centering
    \caption{Backbone ablation on 9 benchmarks. $\Delta$ compares STAL with the DDA baseline using the same spatial detector backbone.}
    \label{tab:backbone_ablation}

    \small
    \setlength{\tabcolsep}{18pt}
    \renewcommand{\arraystretch}{1.05}

    \resizebox{0.9\linewidth}{!}{%
      \begin{tabular}{lcc}
        \toprule
        \textbf{Method} & \textbf{BAL (\%)} & \textbf{$\Delta$} \\
        \midrule
        DDA (DINOv2-L) & 90.0 & -- \\
        \rowcolor[HTML]{E3F1FF}
        STAL (DINOv2-L) & 91.2 & +1.2 \\
        DDA (DINOv3-L) & 93.2 & -- \\
        \rowcolor[HTML]{E3F1FF}
        STAL (DINOv3-L) & 94.8 & +1.6 \\
        DDA* (DINOv3-H+) & 94.2 & -- \\
        \rowcolor[HTML]{E3F1FF}
        \textbf{STAL (DINOv3-H+)}
          & \textbf{97.0}
          & \textbf{+2.8} \\
        \bottomrule
      \end{tabular}%
    }
  \end{table}


\begin{table}[t]
    \centering
    \caption{Loss ablation on 9 benchmarks. $\mathcal{L}_{\text{spatial}}$ supervises the spatial detector, $\mathcal{L}_{\text{align}}$ aligns the projected spatial representation with the frequency
    teacher, and $\mathcal{L}_{\text{tail}}$ and $\mathcal{L}_{\text{freq}}$ supervise the tail head and frequency context of the teacher, respectively. $
    \Delta$ is measured against the spatial-only baseline.}
    \label{tab:loss_ablation}

    \small
    \setlength{\tabcolsep}{10pt}
    \renewcommand{\arraystretch}{1.05}

    \resizebox{0.9\linewidth}{!}{%
      \begin{tabular}{ccccrr}
        \toprule
        $\mathcal{L}_{\text{spatial}}$ &
        $\mathcal{L}_{\text{align}}$ &
        $\mathcal{L}_{\text{tail}}$ &
        $\mathcal{L}_{\text{freq}}$ &
        \textbf{BAL (\%)} &
        \textbf{$\Delta$} \\
        \midrule
        \cmark & \xmark & \xmark & \xmark & 94.2 & -- \\
        \cmark & \cmark & \xmark & \xmark & 94.5 & +0.3 \\
        \cmark & \cmark & \cmark & \xmark & 96.3 & +2.1 \\
        \rowcolor[HTML]{E3F1FF}
        \cmark & \cmark & \cmark & \cmark
          & \textbf{97.0}
          & \textbf{+2.8} \\
        \bottomrule
      \end{tabular}%
    }
  \end{table}

\begin{figure}[t]
\centering
\includegraphics[width=0.9\linewidth]{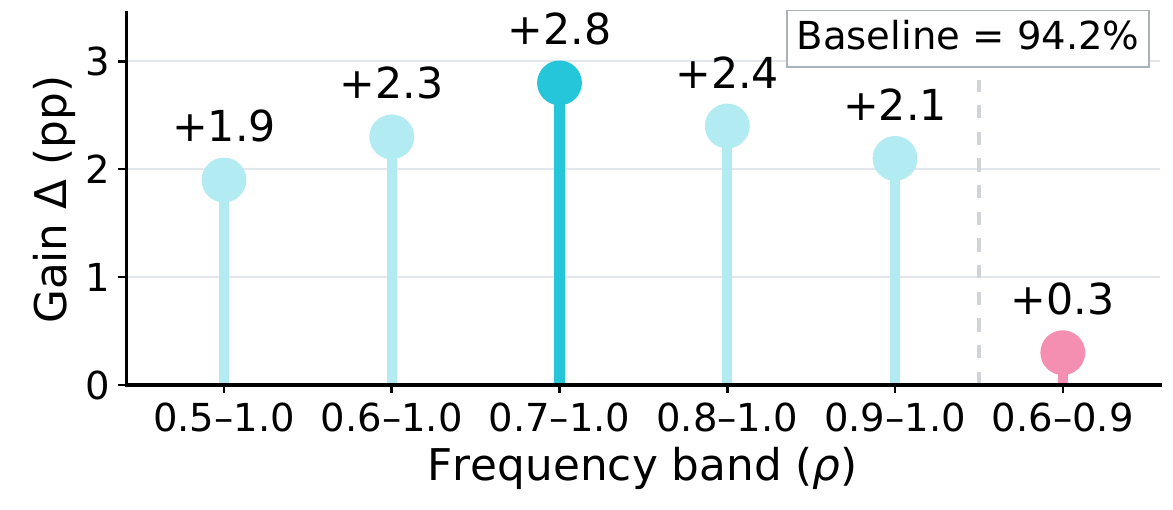}
\caption{Frequency band ablation of the tail head on 9 benchmarks. Each point shows the gain $\Delta$ in BAL over 9 benchmarks relative to the spatial-only baseline (94.2\%) when the tail head takes the radial spectrum on the indicated band $\rho$ as input. The $[0.6, 0.9]$ (pink) denotes a band of the same width that excludes the ultra-high-frequency end of the spectrum.}
\label{fig:band_ablation}
\end{figure}

All ablation results are reported in terms of average balanced accuracy, denoted as BAL, across nine benchmarks.

\noindent\textbf{Frequency Cue Utilization.}
Directly combining the spatial and frequency branches at inference improves BAL to 96.0\%, confirming that spectral cues are discriminative. In contrast, STAL uses the frequency branch solely for training-time auxiliary supervision and achieves 97.0\% BAL, outperforming dual-branch inference by 1.0 percentage point without additional inference overhead.
  
\noindent\textbf{Components of the Frequency Teacher.}
Using frequency context alone yields a marginal gain of 0.2 percentage points, whereas the tail head alone improves BAL by 1.9 percentage points, accounting for the majority of the improvement. Combining both components achieves 97.0\% BAL, with a joint gain of 2.8 percentage points that exceeds the sum of their individual gains of 2.1 percentage points. This superadditive effect highlights the complementary roles of the two components: the tail head captures localized spectral morphology, while the frequency context provides the global spectral trend against which these local patterns are interpreted.

\noindent\textbf{Loss Functions.}
Introducing only $\mathcal{L}_{\mathrm{align}}$ yields a modest improvement of 0.3 percentage points. Adding $\mathcal{L}_{\mathrm{tail}}$ further increases BAL by 1.8 percentage points to 96.3\%, representing the largest incremental gain among the auxiliary loss terms and highlighting the importance of tail-specific supervision. Incorporating $\mathcal{L}_{\mathrm{freq}}$ provides an additional improvement of 0.7 percentage points. 
These results show that explicitly supervising the frequency teacher, particularly its tail head, is crucial for STAL.

\noindent\textbf{Input Frequency Band of the Tail Head.}
All frequency bands that include the spectral endpoint at $\rho=1$ yield improvements of at least 1.9 percentage points. In contrast, the band from $\rho=0.6$ to $\rho=0.9$, which has the same width as the default band but excludes the endpoint, achieves only a 0.3-point gain, substantially below the 2.1-point improvement obtained with the narrower band from $\rho=0.9$ to $\rho=1.0$. This comparison indicates that the improvement stems from the extreme high-frequency tail itself, rather than the additional parameters of the tail head or a wider input frequency range, providing direct evidence that spectral tail uplift constitutes an effective discriminative cue exploited by STAL. Among the bands containing the spectral endpoint, the range from $\rho=0.7$ to $\rho=1.0$ performs best. Extending the lower boundary introduces mid-frequency components dominated by power-law decay, thereby diluting the distinctive tail morphology. Conversely, raising the lower boundary truncates the descending segment preceding the spectral minimum, making it difficult to capture the characteristic decline-then-rise pattern.

\noindent\textbf{Backbone Architectures.}
STAL improves BAL by 1.2, 1.6 and 2.8 \% with DINOv2-L, DINOv3-L and DINOv3-H+, respectively. Notably, the larger improvement is achieved with the stronger backbone, despite its already high baseline BAL of 94.2\%. This suggests that spectral-tail auxiliary supervision remains complementary to improvements in backbone representation capacity.

\section{Conclusion}
\label{conclusion}
We identify spectral tail uplift, a pronounced rise in the ultra-high-frequency spectrum of AI-generated images, and attribute it to nonlinear harmonic accumulation in trained generators. Building on this finding, we propose Spectral Tail Auxiliary Learning (STAL), a generalizable detector that uses spectral-tail cues as auxiliary supervision during training. STAL transfers tail information to the spatial detector without frequency-domain inputs or additional inference branches. Experiments on public benchmarks show strong performance, generalization, and stability across generators and data distributions.

\FloatBarrier
{\small
  \bibliographystyle{ieeenat_fullname}
  \bibliography{references}
}

\clearpage
\appendix
\section*{Appendix}
\addcontentsline{toc}{section}{Appendix}
\section{Controlled Experiments of Spectral Tail Uplift}
\label{app:Controlled_Experiments}
\subsection{Spectral Tail Uplift under JPEG Compression}

\begin{figure*}[!t]
\centering
\includegraphics[width=\linewidth]{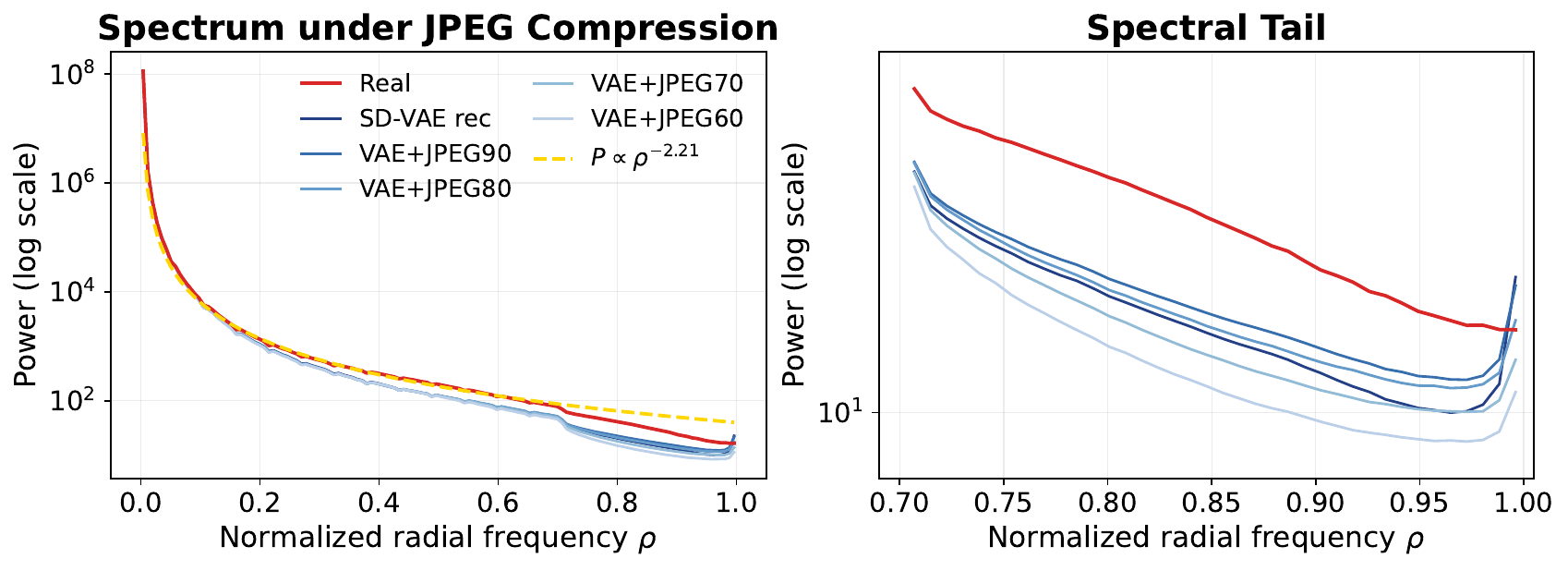}
\caption{Effect of JPEG compression on the spectral tail. We apply JPEG compression with different quality factors to SD-VAE \citep{Rombach2022LDM} reconstructed images and compare their radial FFT power spectra with real images. The yellow curve denotes the power-law decay fitted from the data of real images and serves as a reference. \textbf{Left:} spectra over the full radial frequency range. \textbf{Right:} spectral tail over the local frequency range $\rho \in [0.7,1]$.}
\label{fig:JPEG_analy}
\vspace{-3mm}
\end{figure*}

Due to the loss of high-frequency information caused by JPEG compression, we evaluate how JPEG compression affects spectral tail uplift. We first reconstruct real images with the SD-ema VAE decoder \citep{Rombach2022LDM} to obtain fake samples while reducing semantic bias. We then apply JPEG compression to the SD-VAE reconstructions with quality factors of 90, 80, 70, and 60, and compare them with real images and uncompressed SD-VAE reconstructions. The spectra are computed using the same pipeline as in the main spectral analysis. The yellow dashed curve shows the power-law decay fitted from real-image data as a reference.

As shown in Fig.~\ref{fig:JPEG_analy}, JPEG compression reduces the high-frequency energy of VAE reconstructions, and lower quality factors lead to lower overall energy in the tail region. However, even at JPEG quality 60, the tail uplift in VAE-reconstructed spectra is still not fully suppressed. A clear anomalous upward trend remains visible in the spectral tail. This suggests that spectral tail uplift is not limited to lossless reconstruction settings, but remains an identifiable structural spectral cue after common post-processing, providing useful evidence for generated-image detection.

\subsection{Effects of Nonlinear Activations and Trained Weights}

The spectra above show that the VAE reconstruction process introduces additional high-frequency energy and produces spectral tail uplift in the reconstructed samples. We therefore conduct controlled experiments to investigate how tail uplift arises. We hypothesize that tail uplift is related to harmonic accumulation induced by nonlinear activations in generative models. To verify our hypothesis, we study the effects of nonlinear activations and trained weights on spectral tail uplift. We use both pink-noise inputs and real-image inputs, and reconstruct them with SD-ema VAE, ensuring that each output is paired with its corresponding input to reduce semantic bias.

\paragraph{Effect of nonlinear activations.}
To isolate the effect of nonlinear activations on tail uplift, we keep the trained SD-VAE weights fixed and replace the original SiLU activations with different alternatives. We consider four settings: the original SiLU, Identity, ReLU, and LeakyReLU with negative slope $0.01$. Here, Identity denotes the identity mapping, which is equivalent to completely removing nonlinear activations and reducing the network to a purely linear transformation. As shown in Fig.~\ref{fig:VAE_activation} in the main text, replacing SiLU with Identity reduces the tail uplift to $0.01$ for pink-noise inputs and $0.00$ for real-image inputs, almost eliminating the effect. It indicates that nonlinear activations are crucial for producing spectral tail uplift. In contrast, replacing SiLU with ReLU or LeakyReLU substantially strengthens the uplift, suggesting that different nonlinear activations induce harmonic generation with different strengths. These results suggest that tail uplift is closely related to harmonic accumulation caused by nonlinear activations in generative models.

\begin{figure*}[!t]
\centering
\includegraphics[width=\linewidth]{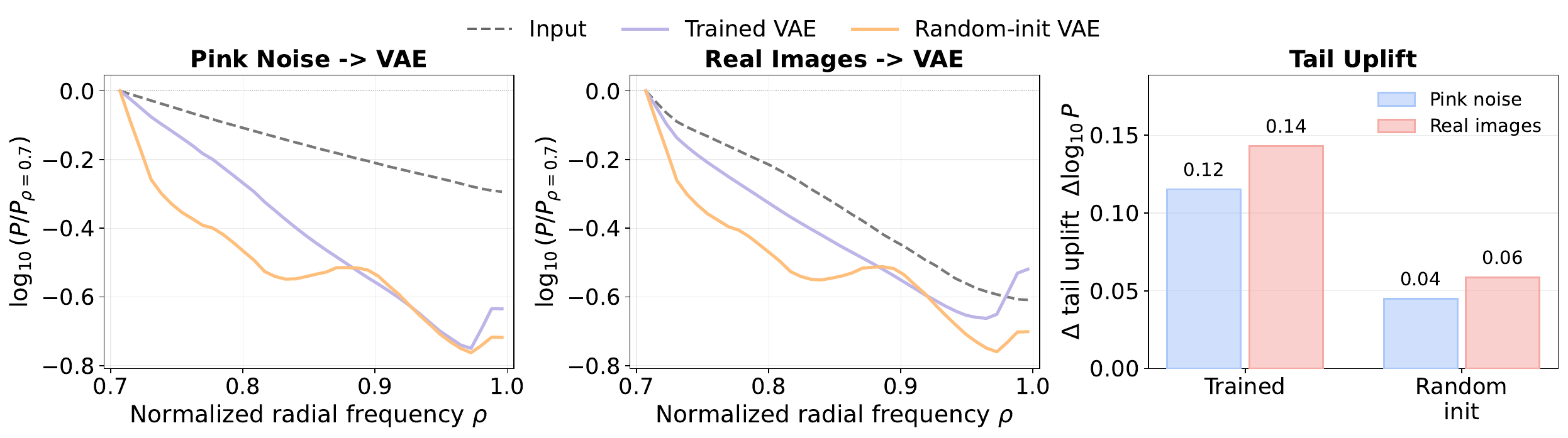}
\caption{Comparison of spectral-tail under trained and random VAE weights. We keep the SD-VAE \citep{Rombach2022LDM} architecture fixed and compare trained weights with random-initialized weights using pink noise (\textbf{left}) and real images (\textbf{middle}) as inputs. Normalized curves show spectra on $\rho\in[0.7,1]$. \textbf{Right:} tail uplift $\Delta\log_{10}P$, the rise from the tail's minimum to $\rho=1$.}
\label{fig:random_init}
\vspace{-3mm}
\end{figure*}

\paragraph{Effect of trained weights.}
We further examine the role of generative-model weights by comparing trained weights with random weights. In the random-weight setting, we load the same SD-VAE architecture but reset all learnable weights using Kaiming random initialization \citep{Kaiming_initialization}. This keeps the architecture, activation functions, depth, channel dimensions, normalization layers, residual connections, and upsampling modules the same as those of the trained SD-VAE. The only difference is whether the weights have undergone generative-model training. As shown in the Fig.~\ref{fig:random_init}, the trained SD-VAE produces tail uplift values of $+0.12$ and $+0.14$ for pink-noise and real-image inputs, respectively. Under random weights, the same architecture produces much weaker uplift values of $+0.04$ and $+0.06$. The pink-noise input itself does not exhibit tail uplift, with an input baseline of $+0.00$. Therefore, the output uplift directly reflects how the network transforms the input spectrum. For pink-noise inputs, the $+0.04$ uplift under random weights can be attributed to weak harmonics generated enerated by nonlinear activations under random initialization. The trained VAE further amplifies this effect to $+0.12$, a three-fold increase. This suggests that training shapes the layer-wise frequency responses of the network, causing harmonic energy to accumulate across layers and become systematically concentrated in the spectral tail. The results on real-image inputs show the same trend.

\begin{figure*}[!t]
\centering
\includegraphics[width=\linewidth]{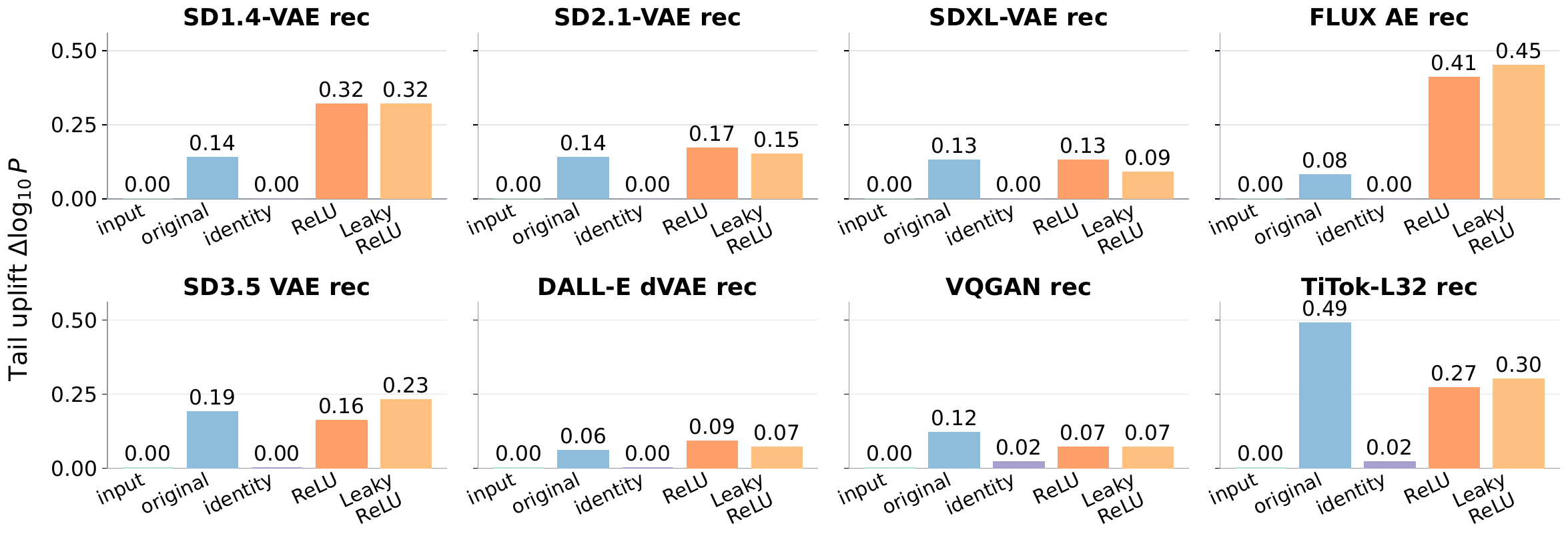}
\caption{Controlled activation-function replacement experiments across eight different VAEs or tokenizers. We replace the original activation fuction with Identity, ReLU, or LeakyReLU, to expose the effect of nonlinear activations.}
\label{fig:reconstruction_bars}
\vspace{-3mm}
\end{figure*}

\begin{figure*}[!t]
\centering
\includegraphics[width=\linewidth]{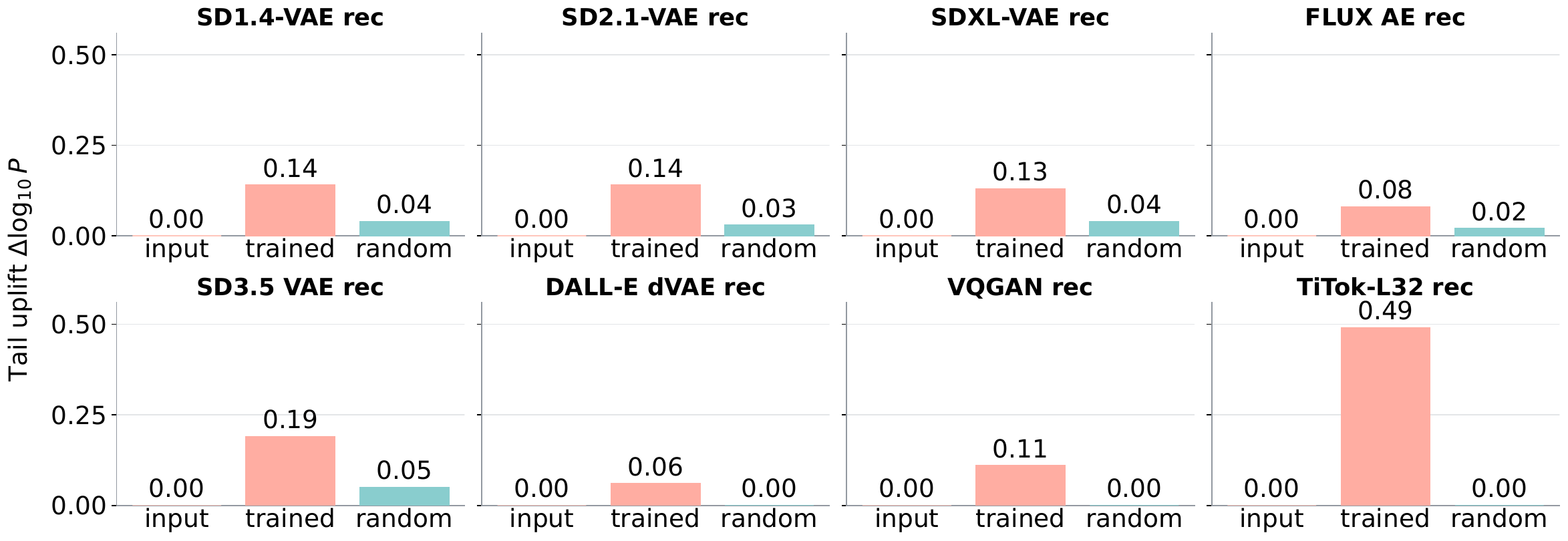}
\caption{Controlled weights experiments across eight different VAEs or tokenizers. We replace the trained weights with random weights.}
\label{fig:trained_random_bars}
\vspace{-3mm}
\end{figure*}

As shown in Fig.~\ref{fig:reconstruction_bars} and Fig.~\ref{fig:trained_random_bars}, we repeat the controlled experiments on activation functions and network weights across eight additional VAEs and tokenizers, and observe consistent results. Replacing the original nonlinear activation functions with the identity function eliminate the spectral tail uplift, while replacing the trained weights with randomly initialized weights substantially suppresses or completely eliminates the tail uplift. Taken as a whole, these controlled experiments indicate that tail uplift depends on two factors. Nonlinear activations provide the ability to generate new frequency components, while trained weights make these components accumulate across layers and align with the spectral tail. Removing either factor substantially weakens or even suppresses tail uplift. With both factors present, the uplift reaches its full strength, removing the nonlinearity eliminates it, and removing training
reduces it to roughly one third. These results establish that both nonlinear activations and trained weights are required for tail uplift, consistent with harmonic accumulation induced by their joint effect.

\section{Formal Analysis of Spectral Tail Uplift}
\label{app:theory}

This appendix provides the formal statements and proofs used in the main text to analyze spectral tail uplift. The analysis is intentionally based on a minimal signal model: a one-dimensional periodic signal, a linear convolutional filter represented by its frequency response, and a pointwise nonlinear activation. This simplified model is not intended to describe every architectural detail of modern generators. Instead, it isolates the mechanism needed by the main text: pointwise nonlinear activations generate new harmonics, and cascaded filters modulate whether these harmonics survive along the decoder.

\subsection{Harmonic Generation by Pointwise Nonlinearity}
\label{app:harmonic_generation}

\noindent\textbf{Theorem 1 (Harmonic generation).}
Let $x:\mathbb{R}\to\mathbb{R}$ be a real-valued $2\pi$-periodic signal with finite bandwidth $M$,
\begin{equation}
  x(t)=\sum_{m=-M}^{M}\hat{x}_m e^{imt},
  \qquad
  \hat{x}_M\neq 0,
  \qquad
  \hat{x}_{-m}=\overline{\hat{x}_m}.
  \label{eq:app_input_signal}
\end{equation}
Let $\phi:\mathbb{R}\to\mathbb{R}$ be a degree-$d$ polynomial activation,
\begin{equation}
  \phi(z)=\sum_{q=0}^{d}a_q z^q,
  \qquad
  a_d\neq 0,\quad d\ge 2.
  \label{eq:app_polynomial_phi}
\end{equation}
Then $\phi(x)$ has highest positive frequency $dM$, and the Fourier coefficient at this frequency is
\begin{equation}
  \widehat{\phi(x)}_{dM}=a_d(\hat{x}_M)^d,
  \qquad
  \bigl|\widehat{\phi(x)}_{dM}\bigr|^2
  =|a_d|^2|\hat{x}_M|^{2d}.
  \label{eq:app_topcoef}
\end{equation}

\noindent\emph{Proof.}
Since $\phi$ is a polynomial, we can write
\begin{equation}
  \phi(x(t))=\sum_{q=0}^{d}a_q[x(t)]^q.
  \label{eq:app_phi_expand}
\end{equation}
For each $q$, the Fourier coefficient of $[x(t)]^q$ at frequency $k$ is the $q$-fold convolution of the Fourier coefficients of $x$:
\begin{equation}
  \widehat{[x]^q}_{k}
  =
  \sum_{\substack{m_1+\cdots+m_q=k\\ |m_j|\le M}}
  \hat{x}_{m_1}\hat{x}_{m_2}\cdots\hat{x}_{m_q}.
  \label{eq:app_qfold_conv}
\end{equation}
Because each $m_j$ lies in $[-M,M]$, the support of $[x]^q$ is contained in $[-qM,qM]$. Therefore, any term with $q<d$ cannot contribute to frequency $dM$.

For $q=d$ and $k=dM$, the constraint $m_1+\cdots+m_d=dM$ with $m_j\le M$ has only one feasible solution: $m_1=\cdots=m_d=M$. Hence
\begin{equation}
  \widehat{[x]^d}_{dM}=(\hat{x}_M)^d.
  \label{eq:app_xd_topcoef}
\end{equation}
Combining this with Eq.~\eqref{eq:app_phi_expand} gives
\begin{equation}
  \widehat{\phi(x)}_{dM}
  =
  \sum_{q=0}^{d}a_q\widehat{[x]^q}_{dM}
  =
  a_d(\hat{x}_M)^d.
  \label{eq:app_theorem1_end}
\end{equation}
Since $a_d\neq 0$ and $\hat{x}_M\neq 0$, this coefficient is nonzero, so the highest positive frequency is exactly $dM$. Taking the squared magnitude gives Eq.~\eqref{eq:app_topcoef}. 

\subsection{Harmonic-Chain Propagation in a Decoder Cascade}
\label{app:harmonic_chain}

\noindent\textbf{Theorem 2 (Harmonic-chain propagation).}
Consider the $L$-layer cascade
\begin{equation}
  y_\ell=H_\ell*x_{\ell-1},
  \qquad
  x_\ell=\phi(y_\ell),
  \qquad
  \ell=1,\ldots,L,
  \label{eq:app_cascade}
\end{equation}
where $H_\ell$ is a linear convolutional filter with frequency response $H_\ell(k)$, and $\phi$ is the degree-$d$ polynomial activation in Eq.~\eqref{eq:app_polynomial_phi}. Let the input be a single-tone signal
\begin{equation}
  x_0(t)=A\cos(k_0t),
  \qquad
  \hat{x}_{0,k_0}=\hat{x}_{0,-k_0}=A/2.
  \label{eq:app_single_tone}
\end{equation}
Assume the filters are non-degenerate along the highest-order harmonic chain, i.e.,
\begin{equation}
  H_\ell(d^{\ell-1}k_0)\neq 0,
  \qquad
  \ell=1,\ldots,L.
  \label{eq:app_nondegenerate}
\end{equation}
Then the output power at the top harmonic $d^Lk_0$ is
\begin{equation}
  \bigl|\hat{x}_L(d^L k_0)\bigr|^2
  =
  |a_d|^{\frac{2(d^L-1)}{d-1}}
  \left(\frac{A}{2}\right)^{2d^L}
  \prod_{\ell=1}^{L}
  \bigl|H_\ell(d^{\ell-1}k_0)\bigr|^{2d^{L-\ell+1}}.
  \label{eq:app_toppower}
\end{equation}

\noindent\emph{Proof.}
We prove the result by induction along the highest-order harmonic chain. A linear convolution does not change the frequency support, in the Fourier domain it only rescales each coefficient:
\begin{equation}
  \hat{y}_\ell(k)=H_\ell(k)\hat{x}_{\ell-1}(k).
  \label{eq:app_filter_response}
\end{equation}

For $L=1$, the pre-activation coefficient at frequency $k_0$ is
\begin{equation}
  \hat{y}_1(k_0)=H_1(k_0)\frac{A}{2}.
  \label{eq:app_base_pre}
\end{equation}
Applying Theorem~1 to $y_1$ gives
\begin{equation}
  \hat{x}_1(dk_0)
  =
  a_d\left(H_1(k_0)\frac{A}{2}\right)^d.
  \label{eq:app_base_coef}
\end{equation}
Thus
\begin{equation}
  |\hat{x}_1(dk_0)|^2
  =
  |a_d|^2|H_1(k_0)|^{2d}\left(\frac{A}{2}\right)^{2d},
  \label{eq:app_base_power}
\end{equation}
which is Eq.~\eqref{eq:app_toppower} when $L=1$.

Now assume that after layer $\ell-1$ the contribution of the highest-order harmonic chain satisfies
\begingroup
\footnotesize
\begin{equation}
\bigl|\hat{x}_{\ell-1}(d^{\ell-1}k_0)\bigr|^2
  =
  |a_d|^{\frac{2(d^{\ell-1}-1)}{d-1}}
  \left(\frac{A}{2}\right)^{2d^{\ell-1}}
  \prod_{j=1}^{\ell-1}
  \bigl|H_j(d^{j-1}k_0)\bigr|^{2d^{\ell-j}}.
  \label{eq:app_induction_hypothesis}
\end{equation}
\endgroup
The next linear filter gives
\begin{equation}
  \bigl|\hat{y}_{\ell}(d^{\ell-1}k_0)\bigr|^2
  =
  \bigl|H_\ell(d^{\ell-1}k_0)\bigr|^2
  \bigl|\hat{x}_{\ell-1}(d^{\ell-1}k_0)\bigr|^2.
  \label{eq:app_induction_filter}
\end{equation}
Along the highest-order chain, the degree-$d$ term contributes to frequency $d^\ell k_0$ by combining the component at $d^{\ell-1}k_0$ exactly $d$ times. Other lower-frequency components cannot reach this top frequency through the degree-$d$ term. Therefore, the chain contribution obeys
\begin{equation}
  \bigl|\hat{x}_{\ell}(d^\ell k_0)\bigr|^2
  =
  |a_d|^2
  \bigl|\hat{y}_{\ell}(d^{\ell-1}k_0)\bigr|^{2d}.
  \label{eq:app_induction_step}
\end{equation}
Substituting Eqs.~\eqref{eq:app_induction_hypothesis} and \eqref{eq:app_induction_filter} into Eq.~\eqref{eq:app_induction_step} yields
\begin{equation}
  \bigl|\hat{x}_{\ell}(d^\ell k_0)\bigr|^2
  =
  |a_d|^{\frac{2(d^\ell-1)}{d-1}}
  \left(\frac{A}{2}\right)^{2d^\ell}
  \prod_{j=1}^{\ell}
  \bigl|H_j(d^{j-1}k_0)\bigr|^{2d^{\ell-j+1}}.
  \label{eq:app_induction_result}
\end{equation}
This is the desired expression for the highest-order chain at layer $\ell$, so the induction is complete. 

\subsection{Scope of the Formal Model}
\label{app:theory_scope}

\begin{table*}[!t]
\renewenvironment{appendixtable}{\par\begingroup\centering\captionsetup{type=table}}{\par\endgroup\medskip}
\begin{appendixtable}
  \centering
  \caption{Comparison on AIGCDetectionBenchmark. Balanced accuracy (\%) is reported.}

  \small
  \setlength{\tabcolsep}{2pt}
  \renewcommand{\arraystretch}{0.98}
  \resizebox{\textwidth}{!}{%
  \begin{tabular}{@{}lcccccccccccccccccc@{}}
  \toprule
  \textbf{Method}
    & \textbf{ADM}
    & \textbf{DALL$\cdot$E 2}
    & \textbf{GLIDE}
    & \textbf{Midjourney}
    & \textbf{VQDM}
    & \textbf{BigGAN}
    & \textbf{CycleGAN}
    & \textbf{GauGAN}
    & \textbf{ProGAN}
    & \textbf{SDXL}
    & \textbf{SD14}
    & \textbf{SD15}
    & \textbf{StarGAN}
    & \textbf{StyleGAN}
    & \textbf{StyleGAN2}
    & \textbf{WFR}
    & \textbf{Wukong}
    & \textbf{Avg} \\
  \midrule

  NPR (CVPR'24)~\cite{Tan2024NPR}
    & 43.8 & 20.0 & 41.2 & 53.4 & 48.4 & 53.1 & 76.6 & 42.2 & 58.7
    & 59.6 & 55.1 & 55.0 & 67.4 & 57.9 & 54.6 & 58.8 & 57.4
    & 53.1 $\pm$ 12.2 \\

  \addlinespace[1pt]
  UnivFD (CVPR'23)~\cite{Ojha2023UnivFD}
    & 62.5 & 50.0 & 61.3 & 55.1 & 76.9 & 87.5 & \underline{96.9}
    & \underline{98.8} & \textbf{99.4}
    & 58.2 & 55.6 & 55.7 & 95.1 & 80.0 & 69.4 & 69.2 & 61.1
    & 72.5 $\pm$ 17.3 \\

  \addlinespace[1pt]
  FatFormer (CVPR'24)~\cite{Liu2024fatformer}
    & 80.2 & 68.5 & 91.1 & 54.4 & 88.0 & \textbf{99.2}
    & \textbf{99.5} & \textbf{99.1} & 98.5
    & 71.7 & 67.5 & 67.2 & \underline{99.4} & \textbf{98.0}
    & \textbf{98.8} & 88.3 & 75.6
    & 85.0 $\pm$ 14.9 \\

  \addlinespace[1pt]
  SAFE (KDD'25)~\cite{Li2025SAFE}
    & 49.5 & 49.5 & 53.0 & 49.0 & 50.2 & 52.2 & 51.9 & 50.0 & 50.0
    & 49.8 & 49.7 & 49.8 & 50.1 & 50.0 & 50.0 & 49.8 & 50.3
    & 50.3 $\pm$ 1.1 \\

  \addlinespace[1pt]
  C2P-CLIP (AAAI'25)~\cite{Tan2025C2PCLIP}
    & 71.6 & 52.3 & 73.5 & 56.6 & 73.7 & \underline{98.4} & 96.8
    & \underline{98.8} & \underline{99.3}
    & 62.3 & 77.5 & 76.9 & \textbf{99.6} & 93.1 & 79.4
    & \underline{94.8} & 79.4
    & 81.4 $\pm$ 15.6 \\

  \addlinespace[1pt]
  AIDE (ICLR'25)~\cite{yan2025AIDE}
    & 52.9 & 51.1 & 60.2 & 49.8 & 69.3 & 70.1 & 93.6 & 60.6 & 89.0
    & 49.6 & 51.6 & 51.0 & 72.1 & 66.5 & 59.0 & 80.6 & 54.5
    & 63.6 $\pm$ 13.9 \\

  \addlinespace[1pt]
  DRCT (ICML'24)~\cite{Chen2024DRCT}
    & 79.9 & 89.2 & 89.2 & 85.5 & 88.6 & 81.4 & 91.0
    & 93.8 & 71.1
    & 88.3 & 91.4 & 91.0 & 53.0 & 62.7 & 63.8 & 73.9 & 90.8
    & 81.4 $\pm$ 12.2 \\

  \addlinespace[1pt]
  AlignedForensics (ICLR'25)~\cite{rajan2025Aligned}
    & 51.6 & 52.0 & 55.6 & 96.2 & 72.1 & 51.2 & 49.5 & 50.8 & 50.7
    & 95.1 & \textbf{99.7} & \textbf{99.6} & 53.8 & 52.7 & 51.6
    & 50.0 & \textbf{99.6}
    & 66.6 $\pm$ 21.6 \\

  \addlinespace[1pt]
  DDA (NeurIPS'25)~\cite{chen2026DDA}
    & 89.5 & 94.6 & 89.6 & 95.6 & 76.6 & 91.0 & 72.5
    & 92.7 & 92.8
    & \underline{99.4} & 98.7 & 98.6 & 72.7 & 87.8 & 90.2 & 52.1 & 98.8
    & 87.8 $\pm$ 12.6 \\

  \midrule
  DDA*
    & \underline{97.0} & \textbf{98.8} & \underline{97.5} & \underline{96.8}
    & \underline{97.9} & 91.3 & 90.8 & 90.9 & 81.8
    & 98.8 & 97.8 & 97.7 & 75.2 & 83.5 & 86.2 & 91.8 & 97.8
    & \underline{92.4 $\pm$ 7.0} \\

  \addlinespace[1pt]
  \rowcolor[HTML]{E3F1FF}
  \textbf{STAL (ours)}
    & \textbf{97.3} & \underline{98.4} & \textbf{98.0} & \textbf{98.2}
    & \textbf{99.4} & 97.7 & 95.9 & 97.8 & 95.3
    & \textbf{99.8} & \underline{99.5} & \underline{99.3} & 81.7
    & \underline{95.9} & \underline{96.7} & \textbf{96.4} & \underline{99.4}
    & \textbf{96.9 $\pm$ 4.2} \\
  \bottomrule
  \end{tabular}%
  }
  \label{tab:aigcdetect}
\end{appendixtable}

\begin{appendixtable}
  \centering
  \caption{Comparison on GenImage. Balanced accuracy (\%) is reported.}
  \small
  \setlength{\tabcolsep}{2pt}
  \renewcommand{\arraystretch}{0.98}
  \resizebox{\textwidth}{!}{%
  \begin{tabular}{@{}lccccccccc@{}}
  \toprule
  \textbf{Method} & \textbf{Midjourney} & \textbf{SDv1.4} & \textbf{SDv1.5} & \textbf{ADM} & \textbf{GLIDE} & \textbf{Wukong} & \textbf{VQDM} & \textbf{BigGAN} & \textbf{Avg} \\
  \midrule
  NPR (CVPR'24)~\cite{Tan2024NPR} & 53.4 & 55.1 & 55.0 & 43.8 & 41.2 & 57.4 & 48.4 & 57.7 & 51.5 $\pm$ 6.3 \\

  \addlinespace[1pt]
  UnivFD (CVPR'23)~\cite{Ojha2023UnivFD} & 55.1 & 55.6 & 55.7 & 62.5 & 61.3 & 61.1 & 76.9 & 84.4 & 64.1 $\pm$ 10.8 \\

  \addlinespace[1pt]
  FatFormer (CVPR'24)~\cite{Liu2024fatformer} & 52.1 & 53.6 & 53.8 & 61.4 & 65.5 & 60.9 & 72.5 & 82.2 & 62.8 $\pm$ 10.4 \\

  \addlinespace[1pt]
  SAFE (KDD'25)~\cite{Li2025SAFE} & 49.0 & 49.7 & 49.8 & 49.5 & 53.0 & 50.3 & 50.2 & 50.9 & 50.3 $\pm$ 1.2 \\

  \addlinespace[1pt]
  C2P-CLIP (AAAI'25)~\cite{Tan2025C2PCLIP} & 56.6 & 77.5 & 76.9 & 71.6 & 73.5 & 79.4 & 73.7 & 85.9 & 74.4 $\pm$ 8.4 \\

  \addlinespace[1pt]
  AIDE (ICLR'25)~\cite{yan2025AIDE} & 58.2 & 77.2 & 77.4 & 50.4 & 54.6 & 70.5 & 50.8 & 50.6 & 61.2 $\pm$ 11.9 \\

  \addlinespace[1pt]
  DRCT (ICML'24)~\cite{Chen2024DRCT} & 82.4 & 88.3 & 88.2 & 76.9 & 86.1 & 87.9 & 85.4 & 87.0 & 84.7 $\pm$ 2.7 \\

  \addlinespace[1pt]
  AlignedForensics (ICLR'25)~\cite{rajan2025Aligned} & \underline{97.5} & \textbf{99.7} & \textbf{99.6} & 52.4 & 57.6 & \textbf{99.6} & 75.0 & 50.6 & 79.0 $\pm$ 22.7 \\

  \addlinespace[1pt]
  DDA (NeurIPS'25)~\cite{chen2026DDA} & 95.6 & 98.7 & 98.6 & 89.5 & 89.6 & 98.7 & 76.5 & 86.5 & 91.7 $\pm$ 7.8 \\

  \midrule
  DDA* & 96.8 & 97.8 & 97.7 & \underline{97.0} & \underline{97.5} & 97.8 & \underline{97.9} & \underline{97.0} & \underline{97.4 $\pm$ 0.4} \\

  \addlinespace[1pt]
  \rowcolor[HTML]{E3F1FF}
  \textbf{STAL (ours)} & \textbf{98.2} & \underline{99.5} & \underline{99.3} & \textbf{97.3} & \textbf{98.0} & \underline{99.4} & \textbf{99.4} & \textbf{97.9} & \textbf{98.6 $\pm$ 0.9} \\
  \bottomrule
  \end{tabular}%
  }
  \label{tab:genimage}
\end{appendixtable}

\end{table*}

\begin{table*}[!t]
\renewenvironment{appendixtable}{\par\begingroup\centering\captionsetup{type=table}}{\par\endgroup\medskip}
\begin{appendixtable}
  \centering
  \caption{Comparison on EvalGEN. Balanced accuracy (\%) is reported.}
  \small
  \setlength{\tabcolsep}{6pt}
  \renewcommand{\arraystretch}{0.98}
  
  \stalFullWidthTable{14}{lcccccc}{%
  \toprule
  \textbf{Method} & \textbf{Flux} & \textbf{GoT} & \textbf{Infinity} & \textbf{NOVA} & \textbf{OmniGen} & \textbf{Avg} \\
  \midrule
  NPR (CVPR'24)~\cite{Tan2024NPR} & 0.7 & 0.2 & 6.5 & 4.7 & 2.2 & 2.9 $\pm$ 2.7 \\

  \addlinespace[1pt]
  UnivFD (CVPR'23)~\cite{Ojha2023UnivFD} & 4.0 & 9.2 & 15.7 & 8.3 & 39.6 & 15.4 $\pm$ 14.2 \\

  \addlinespace[1pt]
  FatFormer (CVPR'24)~\cite{Liu2024fatformer} & 9.9 & 47.9 & 44.7 & 98.3 & 27.3 & 45.6 $\pm$ 33.1 \\

  \addlinespace[1pt]
  SAFE (KDD'25)~\cite{Li2025SAFE} & 1.0 & 0.5 & 1.9 & 0.6 & 1.6 & 1.1 $\pm$ 0.6 \\

  \addlinespace[1pt]
  C2P-CLIP (AAAI'25)~\cite{Tan2025C2PCLIP} & 8.7 & 49.6 & 35.3 & 86.4 & 14.5 & 38.9 $\pm$ 31.2 \\

  \addlinespace[1pt]
  AIDE (ICLR'25)~\cite{yan2025AIDE} & 17.9 & 24.7 & 3.4 & 16.3 & 33.4 & 19.1 $\pm$ 11.1 \\

  \addlinespace[1pt]
  DRCT (ICML'24)~\cite{Chen2024DRCT} & 72.5 & 81.4 & 77.9 & 84.6 & 72.5 & 77.8 $\pm$ 5.4 \\

  \addlinespace[1pt]
  AlignedForensics (ICLR'25)~\cite{rajan2025Aligned} & 32.0 & 72.3 & 74.0 & 84.8 & 77.0 & 68.0 $\pm$ 20.7 \\

  \addlinespace[1pt]
  DDA (NeurIPS'25)~\cite{chen2026DDA} & 89.9 & \underline{99.5} & 97.8 & 99.5 & \underline{99.5} & 97.2 $\pm$ 4.2 \\

  \midrule
  DDA* & \textbf{97.9} & \textbf{99.6} & \underline{99.8} & \underline{99.8} & \textbf{99.7} & \textbf{99.3 $\pm$ 0.8} \\

  \addlinespace[1pt]
  \rowcolor[HTML]{E3F1FF}
  \textbf{STAL (ours)} & \underline{95.7} & \underline{99.5} & \textbf{99.9} & \textbf{99.9} & \underline{99.5} & \underline{98.9 $\pm$ 1.8} \\
  \bottomrule
  }
  \label{tab:evalgen}
\end{appendixtable}

\captionsetup{type=figure}
\centering
\includegraphics[width=\linewidth]{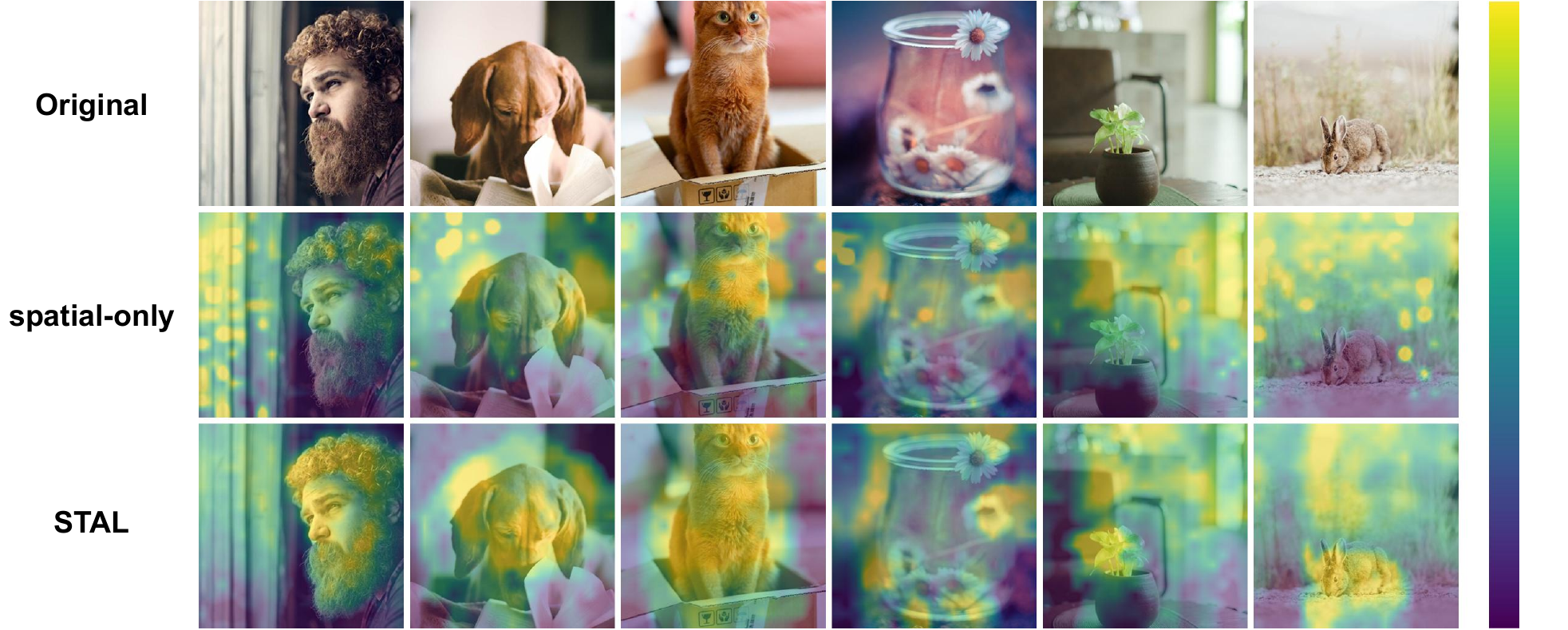}
\caption{Visualization of model attention. From top to bottom, the three rows show the original images, heatmaps generated by the spatial-only model, and heatmaps generated by STAL, respectively. Colors from yellow to purple indicate attention values from high to low.}
\label{fig:heatmap}
\end{table*}

Theorems~1 and~2 are used as a mechanism-level explanation rather than a complete generative model. Several points are important for avoiding over-interpretation.

Theorem~1 is stated for polynomial activations because this setting gives an exact closed-form coefficient. For smooth activations used in modern decoders, a finite Taylor approximation on the bounded activation range leads to the same type of local harmonic-generation mechanism. For piecewise linear activations, harmonics can also be introduced when the input crosses nonlinear kink regions, although the exact coefficients differ. We therefore use the polynomial model as an analytically tractable abstraction of the pointwise activation layer.

Theorem~2 does not claim that arbitrary filters necessarily amplify the harmonic chain. It shows where the filter gains enter the top-harmonic power. If a gain $H_\ell(d^{\ell-1}k_0)$ is close to zero, the corresponding harmonic path is suppressed, if the learned filters preserve or amplify these frequencies, the path can survive to the spectral tail. Thus, convolutional weights should be interpreted as a modulation and propagation factor rather than as an unconditional source of high-frequency energy.

\paragraph{Connection to controlled observations.}
The formal results above identify a mechanism by which decoder pointwise nonlinear activations can create harmonics and by which convolutional filters can modulate their propagation. This mechanism is supported by our controlled SD-VAE experiments. Under the same measurement protocol, real images follow the expected decaying spectral profile and do not exhibit a tail-uplift pattern. After SD-VAE reconstruction, the reconstructed fake images show a clear uplift shape at the spectral tail. To isolate the role of pointwise activations, we fix the trained VAE weights and replace all SiLU activation functions with identity mappings. The tail uplift is then almost completely suppressed, while replacing SiLU with other nonlinear activations restores or strengthens the effect. These observations support our interpretation that pointwise activations inject additional harmonics, while learned convolutional filters modulate how these harmonics propagate through the decoder.

\section{Implementation Details}
\label{app:implementation}

\subsection{Training Data and View Construction}

During training, we use real images from MSCOCO \citep{MSCOCO} and their reconstructed counterparts to construct training data. For the spatial branch, we follow DDA \citep{chen2026DDA} to build aligned training pairs, mitigating reconstruction-induced low-level pixel biases and encouraging the detector to learn more generalizable cues. Different from the spatial branch, the auxiliary frequency branch uses a frequency-preserving view. This view shares the necessary geometric transformations with the spatial view, but avoids augmentations that substantially alter the spectral shape, such as strong compression, resampling, or other operations that may distort spectral-tail cues. This design allows the training-time frequency teacher to capture spectral tail uplift consistently, rather than learning spectral perturbations caused by data augmentation.

\subsection{More Training Details}

We train STAL for 1 epoch on 8 NVIDIA V100 32GB GPUs with a batch size of 16 per GPU. The spatial detector uses DINOv3 ViT-H+/16 \citep{Simeoni2025DINOv3} with LoRA \citep{hu2022lora} fine-tuning, where the input resolution is $336\times336$, the LoRA rank is 8, and $\alpha=1.0$. We use AdamW \citep{Adamw} with a learning rate of $1\times10^{-4}$ and a weight decay of 0.005. The frequency-related losses follow a curriculum schedule and are discarded together with the frequency branch at inference time, so STAL introduces no additional inference cost. The frequency-related losses follow a lightweight curriculum schedule: the weight is linearly warmed up in the first 800 steps, kept at full strength in early training, and then cosine-annealed to zero between 15\% and 45\% of the estimated training steps. This encourages the spatial detector to absorb spectral-tail cues during training while avoiding dependence on the auxiliary frequency branch.

\subsection{Evaluation Protocol}

All experiments report balanced accuracy, defined as the average of real-image accuracy and generated-image accuracy to mitigate the effect of class imbalance. For datasets with multiple subsets, we report the arithmetic mean and standard deviation of balanced accuracy across subsets. All methods \citep{Tan2024NPR, Ojha2023UnivFD, Liu2024fatformer, Li2025SAFE, Tan2025C2PCLIP, yan2025AIDE, Chen2024DRCT, rajan2025Aligned, chen2026DDA}are evaluated using their official model weights. DDA* denotes a DDA variant using the same DINOv3-H+ \citep{Simeoni2025DINOv3} backbone as STAL, which provides a controlled comparison under matched backbone. DDA* and our method are trained under the same setting with fixed random seeds, and the best checkpoint is selected for evaluation to ensure fairness.

\begin{table*}[!t]
\renewenvironment{appendixtable}{\par\begingroup\centering\captionsetup{type=table}}{\par\endgroup\medskip}
\begin{appendixtable}
  \centering
  \caption{Comparison on Synthbuster. Balanced accuracy (\%) is reported.}
  \small
  \setlength{\tabcolsep}{2pt}
  \renewcommand{\arraystretch}{0.98}
  \resizebox{\textwidth}{!}{%
  \begin{tabular}{@{}lcccccccccc@{}}
  \toprule
  \textbf{Method} & \textbf{DALL$\cdot$E 2} & \textbf{DALL$\cdot$E 3} & \textbf{Firefly} & \textbf{GLIDE} & \textbf{Midjourney} & \textbf{SD 1.3} & \textbf{SD 1.4} & \textbf{SD 2} & \textbf{SDXL} & \textbf{Avg} \\
  \midrule
  NPR (CVPR'24)~\cite{Tan2024NPR} & 51.1 & 49.3 & 46.5 & 48.5 & 52.8 & 51.4 & 51.8 & 46.0 & 52.8 & 50.0 $\pm$ 2.6 \\

  \addlinespace[1pt]
  UnivFD (CVPR'23)~\cite{Ojha2023UnivFD} & 83.5 & 47.4 & 89.9 & 53.3 & 52.5 & 70.4 & 69.9 & 75.7 & 68.0 & 67.8 $\pm$ 14.4 \\

  \addlinespace[1pt]
  FatFormer (CVPR'24)~\cite{Liu2024fatformer} & 59.4 & 39.5 & 60.3 & 72.7 & 44.4 & 53.7 & 54.0 & 52.3 & 69.1 & 56.1 $\pm$ 10.7 \\

  \addlinespace[1pt]
  SAFE (KDD'25)~\cite{Li2025SAFE} & 58.0 & 9.9 & 10.3 & 52.2 & 56.7 & 59.4 & 59.1 & 53.0 & 59.5 & 46.5 $\pm$ 20.8 \\

  \addlinespace[1pt]
  C2P-CLIP (AAAI'25)~\cite{Tan2025C2PCLIP} & 55.6 & 63.2 & 59.5 & 86.7 & 52.9 & 75.2 & 76.7 & 69.2 & 77.7 & 68.5 $\pm$ 11.4 \\

  \addlinespace[1pt]
  AIDE (ICLR'25)~\cite{yan2025AIDE} & 34.9 & 33.7 & 24.8 & 65.0 & 57.5 & 74.1 & 73.7 & 53.2 & 68.4 & 53.9 $\pm$ 18.6 \\

  \addlinespace[1pt]
  DRCT (ICML'24)~\cite{Chen2024DRCT} & 77.2 & 86.6 & 84.1 & 82.6 & 73.7 & 86.6 & 86.6 & 83.2 & 71.3 & 84.8 $\pm$ 3.6 \\

  \addlinespace[1pt]
  AlignedForensics (ICLR'25)~\cite{rajan2025Aligned} & 50.2 & 48.9 & 51.7 & 53.5 & \underline{98.7} & \underline{98.8} & \underline{98.8} & \underline{98.6} & \underline{97.3} & 77.4 $\pm$ 25.0 \\

  \addlinespace[1pt]
  DDA (NeurIPS'25)~\cite{chen2026DDA} & 86.3 & \underline{90.0} & 91.9 & 76.5 & 93.5 & 92.9 & 92.7 & 93.3 & 93.5 & 90.1 $\pm$ 5.6 \\

  \midrule
  DDA* & \textbf{99.8} & \textbf{99.9} & \textbf{99.9} & \textbf{99.5} & \textbf{99.9} & \textbf{99.9} & \textbf{99.9} & \textbf{99.9} & \textbf{99.9} & \textbf{99.8 $\pm$ 0.1} \\

  \addlinespace[1pt]
  \rowcolor[HTML]{E3F1FF}
  \textbf{STAL (ours)} & \underline{99.6} & \textbf{99.9} & \underline{99.8} & \underline{98.6} & \textbf{99.9} & \textbf{99.9} & \textbf{99.9} & \textbf{99.9} & \textbf{99.9} & \underline{99.7 $\pm$ 0.4} \\
  \bottomrule
  \end{tabular}%
  }
  \label{tab:synthbuster}
\end{appendixtable}

\end{table*}

\section{More Experiment Results}
\label{app:experiment}
\subsection{More Comparison Results}

In this section, we provide complete detailed results on the benchmarks \citep{Zhu2023GenImage, Chen2024DRCT, Synthbuster, chen2026DDA, Wang2020CNNspot, Zhong2023PatchCraft, SynthWildx, WildRF}
that are not fully reported in the main text. Table~\ref{tab:aigcdetect} reports results on AIGCDetectBenchmark, covering various GAN-based and diffusion-based generators. It provides a comprehensive evaluation of detector generalization across different generator architectures. STAL achieves the best overall performance with an accuracy of 96.9\%, outperforming the second-best method by 4.5\%. In addition, Table~\ref{tab:synthwildx_wildrf} reports the results on two in-the-wild datasets, SynthWildx and WildRF. STAL achieves a balanced accuracy of 95.2\% on SynthWildx and 97.9\% on WildRF. These results indicate that STAL also achieves strong performance in real-world scenarios. Tables~\ref{tab:genimage} and~\ref{tab:drct2m} present detailed results on GenImage and DRCT-2M, respectively, covering a wide range of generative models and variants, including GAN- and diffusion-based generators. Table~\ref{tab:synthbuster} further evaluates STAL on Synthbuster, which includes images from diverse commercial and open-source generative models. Table~\ref{tab:forensynths} reports results on ForenSynths, focusing on generalization to early GAN-based generators and image synthesis methods. Finally, Table~\ref{tab:evalgen} reports detection performance on EvalGEN, which contains recent generative models. Across these detailed benchmarks, STAL achieves the best or near-best average performance in most cases. These results suggest that the proposed auxiliary frequency supervision improves cross-generator generalization across both classical and recent generative architectures.

\subsection{More Robustness Analysis}

In addition to cross-dataset generalization, we further evaluate the stability of STAL under common image post-processing operations. All experiments are conducted on GenImage \citep{Zhu2023GenImage}, where we apply three types of perturbations, including JPEG compression, resizing, and Gaussian blur, and report balanced accuracy. Fig.~\ref{fig:robustness} in the main text shows the robustness results under these three post-processing operations.

\paragraph{JPEG compression.}

Under JPEG compression, when the quality factor decreases from $Q=100$ to $Q=60$, the balanced accuracy of STAL drops from 98.62\% to 96.24\%, corresponding to only a 2.38 percentage point decrease. More importantly, as the compression becomes stronger, the advantage of STAL over the second-best method does not shrink: STAL leads by 3.04 percentage points at $Q=100$ and by 7.73 percentage points at $Q=60$. This shows that STAL performs better not only on uncompressed or mildly compressed images, but also under stronger JPEG recompression.

\paragraph{Resizing.}

For resizing, using $\alpha=1.0$ as the reference point, STAL still reaches 97.27\% at $\alpha=0.5$, with a decrease of 1.35 percentage points. At $\alpha=0.75$, $1.5$, and $2.0$, it obtains 98.36\%, 98.52\%, and 98.50\%, respectively, which are nearly unchanged from the result at $\alpha=1.0$. Across all resize ratios, STAL outperforms the second-best method by at least 3.11 percentage points. Under the most challenging strong downsampling setting, i.e., $\alpha=0.5$, the margin further increases to 9.84 percentage points.

\paragraph{Gaussian blur.}

Under Gaussian blur, blur is the most challenging perturbation for STAL among the three types in terms of performance drop, but the maximum decrease remains within 3.10 percentage points. In terms of comparison with other methods, STAL ranks first at all blur strengths and exceeds the second-best method by 10.73 percentage points at $\sigma=2.0$.

Overall, STAL exhibits the strongest robustness across all perturbation settings, and its advantage becomes larger under more challenging conditions such as strong JPEG compression, strong downsampling, and strong blur. These results indicate that using frequency-domain information as auxiliary supervision for the spatial detector improves cross-generator generalization while maintaining strong adaptability to common image post-processing operations.

\subsection{Heatmap Visualization}

As shown in Fig.~\ref{fig:heatmap}, we visualize the attention maps of the spatial-only baseline and STAL. The spatial-only model, trained without frequency-domain auxiliary supervision, exhibits relatively diffuse attention responses. In contrast, STAL yields more localized responses over regions associated with high-frequency cues, suggesting that the auxiliary frequency supervision encourages the spatial encoder to attend more selectively to frequency-relevant structures.

\begin{table*}[!t]
\renewenvironment{appendixtable}{\par\begingroup\centering\captionsetup{type=table}}{\par\endgroup\medskip}
\begin{appendixtable}
  \centering
  \caption{Comparison on SynthWildx and WildRF. Balanced accuracy (\%) is reported.}
  \small
  \setlength{\tabcolsep}{2pt}
  \renewcommand{\arraystretch}{1.05}
  \resizebox{\textwidth}{!}{%
  \begin{tabular}{@{}lcccccccc@{}}
  \toprule
  \multirow{2}{*}{\textbf{Method}}
    & \multicolumn{4}{c}{\textbf{SynthWildx}}
    & \multicolumn{4}{c@{}}{\textbf{WildRF}} \\
  \cmidrule(lr){2-5} \cmidrule(lr){6-9}
    & \textbf{DALL$\cdot$E 3}
    & \textbf{Firefly}
    & \textbf{Midjourney}
    & \textbf{Avg.}
    & \textbf{Facebook}
    & \textbf{Reddit}
    & \textbf{Twitter}
    & \textbf{Avg.} \\
  \midrule
  NPR (CVPR'24)~\cite{Tan2024NPR} & 43.6 & 61.3 & 44.5 & 49.8 $\pm$ 10.0 & 78.1 & 61.0 & 51.3 & 63.5 $\pm$ 13.6 \\
  UnivFD (CVPR'23)~\cite{Ojha2023UnivFD} & 45.4 & 65.3 & 46.2 & 52.3 $\pm$ 11.3 & 49.1 & 60.2 & 56.5 & 55.3 $\pm$ 5.7 \\
  FatFormer (CVPR'24)~\cite{Liu2024fatformer} & 46.5 & 61.6 & 48.3 & 52.1 $\pm$ 8.2 & 54.1 & 68.1 & 54.4 & 58.9 $\pm$ 8.0 \\
  SAFE (KDD'25)~\cite{Li2025SAFE} & 49.4 & 48.2 & 49.6 & 49.1 $\pm$ 0.7 & 50.9 & 74.1 & 37.5 & 57.2 $\pm$ 18.5 \\
  C2P-CLIP (AAAI'25)~\cite{Tan2025C2PCLIP} & 56.9 & 61.4 & 53.0 & 57.1 $\pm$ 4.2 & 54.4 & 68.4 & 55.9 & 59.6 $\pm$ 7.7 \\
  AIDE (ICLR'25)~\cite{yan2025AIDE} & 63.4 & 48.8 & 51.9 & 48.8 $\pm$ 0.8 & 57.8 & 71.5 & 45.8 & 58.4 $\pm$ 12.9 \\
  DRCT (ICML'24)~\cite{Chen2024DRCT} & 58.3 & 56.4 & 50.5 & 55.1 $\pm$ 1.8 & 46.6 & 53.1 & 55.2 & 50.6 $\pm$ 3.5 \\
  AlignedForensics (ICLR'25)~\cite{rajan2025Aligned} & 85.5 & 58.5 & 92.2 & 78.8 $\pm$ 17.8 & 89.4 & 69.1 & 81.8 & 80.1 $\pm$ 10.3 \\
  DDA (NeurIPS'25)~\cite{chen2026DDA} & \underline{92.3} & 87.3 & \underline{93.1} & 90.9 $\pm$ 3.1 & 93.1 & 86.4 & 91.5 & 90.3 $\pm$ 3.5 \\

  \midrule
  DDA* & \underline{92.3} & \underline{91.7} & 92.0 & \underline{92.0 $\pm$ 0.3} & \underline{96.6} & \underline{95.5} & \underline{96.9} & \underline{96.3 $\pm$ 0.7} \\
  \rowcolor[HTML]{E3F1FF}
  \textbf{STAL (ours)} & \textbf{95.6} & \textbf{94.8} & \textbf{95.4} & \textbf{95.2 $\pm$ 0.4} & \textbf{98.4} & \textbf{96.7} & \textbf{98.6} & \textbf{97.9 $\pm$ 1.0} \\
  \bottomrule
  \end{tabular}%
  }
  \label{tab:synthwildx_wildrf}
\end{appendixtable}

\begin{appendixtable}
  \centering
  \caption{Comparison on DRCT-2M. Balanced accuracy (\%) is reported.}
  \small
  \setlength{\tabcolsep}{2pt}
  \renewcommand{\arraystretch}{0.98}
  \resizebox{\textwidth}{!}{%
  \begin{tabular}{@{}lcccccccccccccc@{}}
  \toprule
  \textbf{Method} & \textbf{LDM} & \textbf{SDv1.4} & \textbf{SDv1.5} & \textbf{SDv2} & \textbf{SDXL} & \textbf{SDXL-Refiner} & \textbf{SD-Turbo} & \textbf{SDXL-Turbo} & \textbf{LCM-SDv1.5} & \textbf{LCM-SDXL} & \textbf{SDv1-Ctrl} & \textbf{SDv2-Ctrl} & \textbf{SDXL-Ctrl} & \textbf{Avg} \\
  \midrule
  NPR (CVPR'24)~\cite{Tan2024NPR} & 33.0 & 29.1 & 29.0 & 35.1 & 33.2 & 28.4 & 27.9 & 27.9 & 29.4 & 30.2 & 28.4 & 28.3 & 34.7 & 30.4 $\pm$ 2.7 \\

  \addlinespace[1pt]
  UnivFD (CVPR'23)~\cite{Ojha2023UnivFD} & 85.4 & 56.8 & 56.4 & 58.2 & 63.2 & 55.0 & 56.5 & 53.0 & 54.5 & 65.9 & 68.0 & 65.4 & 75.9 & 62.6 $\pm$ 9.5 \\

  \addlinespace[1pt]
  FatFormer (CVPR'24)~\cite{Liu2024fatformer} & 55.9 & 48.2 & 48.2 & 48.2 & 48.2 & 48.3 & 48.2 & 48.2 & 48.3 & 50.6 & 49.7 & 49.9 & 59.8 & 50.1 $\pm$ 3.6 \\

  \addlinespace[1pt]
  SAFE (KDD'25)~\cite{Li2025SAFE} & 50.3 & 50.1 & 50.0 & 50.0 & 49.9 & 50.1 & 50.0 & 50.0 & 50.1 & 50.0 & 49.9 & 50.0 & 54.7 & 50.4 $\pm$ 1.3 \\

  \addlinespace[1pt]
  C2P-CLIP (AAAI'25)~\cite{Tan2025C2PCLIP} & 83.0 & 51.7 & 51.7 & 52.9 & 51.9 & 64.6 & 51.7 & 50.6 & 52.0 & 66.1 & 56.9 & 54.7 & 77.8 & 58.9 $\pm$ 10.8 \\

  \addlinespace[1pt]
  AIDE (ICLR'25)~\cite{yan2025AIDE} & 64.4 & 74.9 & 75.1 & 58.5 & 53.5 & 66.3 & 52.8 & 52.8 & 70.0 & 54.3 & 65.9 & 53.6 & 53.9 & 61.2 $\pm$ 8.6 \\

  \addlinespace[1pt]
  DRCT (ICML'24)~\cite{Chen2024DRCT} & 96.7 & 96.3 & 96.3 & 94.9 & 96.2 & 93.5 & 93.4 & 92.9 & 91.2 & 95.0 & 95.6 & 92.7 & 92.0 & 94.4 $\pm$ 1.8 \\

  \addlinespace[1pt]
  AlignedForensics (ICLR'25)~\cite{rajan2025Aligned} & \textbf{99.9} & \textbf{99.9} & \textbf{99.9} & 99.6 & 90.2 & 81.3 & 99.7 & 89.4 & 99.7 & 90.0 & \textbf{99.9} & 99.2 & 87.6 & 95.1 $\pm$ 6.5 \\

  \addlinespace[1pt]
  DDA (NeurIPS'25)~\cite{chen2026DDA} & 99.2 & 98.9 & 99.0 & 98.3 & 98.0 & \underline{96.8} & 97.9 & 94.8 & 95.9 & 98.2 & 98.7 & 99.0 & 99.4 & 98.0 $\pm$ 1.4 \\

  \midrule
  DDA* & \underline{99.8} & 99.5 & 99.5 & \underline{99.7} & \underline{99.7} & \textbf{99.7} & \underline{99.8} & \underline{99.8} & \underline{99.8} & \underline{99.8} & \underline{99.8} & \underline{99.8} & \underline{99.8} & \underline{99.7 $\pm$ 0.1} \\

  \addlinespace[1pt]
  \rowcolor[HTML]{E3F1FF}
  \textbf{STAL (ours)} & \textbf{99.9} & \underline{99.7} & \underline{99.6} & \textbf{99.8} & \textbf{99.8} & \textbf{99.7} & \textbf{99.9} & \textbf{99.9} & \textbf{99.9} & \textbf{99.9} & \textbf{99.9} & \textbf{99.9} & \textbf{99.9} & \textbf{99.8 $\pm$ 0.1} \\
  \bottomrule
  \end{tabular}%
  }
  \label{tab:drct2m}
\end{appendixtable}

\begin{appendixtable}
  \centering
  \caption{Comparison on ForenSynths. Balanced accuracy (\%) is reported.}
  \small
  \setlength{\tabcolsep}{2pt}
  \renewcommand{\arraystretch}{0.98}
  \resizebox{\textwidth}{!}{%
  \begin{tabular}{@{}lcccccccccccccc@{}}
  \toprule
  \textbf{Method} & \textbf{BigGAN} & \textbf{CRN} & \textbf{CycleGAN} & \textbf{DeepFake} & \textbf{GauGAN} & \textbf{IMLE} & \textbf{ProGAN} & \textbf{SAN} & \textbf{SeeingDark} & \textbf{StarGAN} & \textbf{StyleGAN} & \textbf{StyleGAN2} & \textbf{WFR} & \textbf{Avg} \\
  \midrule
  NPR (CVPR'24)~\cite{Tan2024NPR} & 53.1 & 0.4 & 76.6 & 35.7 & 42.2 & 5.3 & 58.7 & 48.4 & 63.6 & 67.4 & 57.9 & 54.6 & 58.8 & 47.9 $\pm$ 22.6 \\

  \addlinespace[1pt]
  UnivFD (CVPR'23)~\cite{Ojha2023UnivFD} & 87.5 & 55.7 & \underline{96.9} & 69.4 & \underline{98.8} & 68.1 & \textbf{99.4} & 58.2 & 62.2 & 95.1 & 80.0 & 69.4 & 69.2 & 77.7 $\pm$ 16.1 \\

  \addlinespace[1pt]
  FatFormer (CVPR'24)~\cite{Liu2024fatformer} & \textbf{99.3} & 72.1 & \textbf{99.5} & \textbf{93.0} & \textbf{99.3} & 72.1 & 98.4 & 70.8 & 81.9 & \underline{99.4} & \textbf{98.1} & \textbf{98.9} & 88.3 & 90.0 $\pm$ 11.8 \\

  \addlinespace[1pt]
  SAFE (KDD'25)~\cite{Li2025SAFE} & 52.2 & 50.0 & 51.9 & 50.1 & 50.0 & 50.0 & 50.0 & 50.9 & 41.1 & 50.1 & 50.0 & 50.0 & 49.8 & 49.7 $\pm$ 2.7 \\

  \addlinespace[1pt]
  C2P-CLIP (AAAI'25)~\cite{Tan2025C2PCLIP} & \underline{98.4} & \underline{93.3} & 96.8 & \underline{92.6} & \underline{98.8} & \underline{93.2} & \underline{99.3} & 63.2 & \textbf{94.7} & \textbf{99.6} & 93.1 & 79.4 & \underline{94.8} & \underline{92.0 $\pm$ 10.1} \\

  \addlinespace[1pt]
  AIDE (ICLR'25)~\cite{yan2025AIDE} & 70.1 & 12.2 & 93.6 & 53.2 & 60.6 & 15.9 & 89.0 & 55.3 & 44.2 & 72.1 & 66.5 & 59.0 & 80.6 & 59.4 $\pm$ 24.6 \\

  \addlinespace[1pt]
  DRCT (ICML'24)~\cite{Chen2024DRCT} & 81.4 & 78.4 & 91.0 & 51.5 & 93.8 & 82.6 & 71.1 & 84.9 & 72.2 & 53.0 & 62.7 & 63.8 & 73.9 & 73.9 $\pm$ 13.4 \\

  \addlinespace[1pt]
  AlignedForensics (ICLR'25)~\cite{rajan2025Aligned} & 51.2 & 50.4 & 49.5 & 71.7 & 50.8 & 49.7 & 50.7 & 67.6 & 51.4 & 53.8 & 52.7 & 51.6 & 50.0 & 53.9 $\pm$ 7.1 \\

  \addlinespace[1pt]
  DDA (NeurIPS'25)~\cite{chen2026DDA} & 91.0 & 87.0 & 72.5 & 76.5 & 92.7 & 89.7 & 92.8 & \underline{94.7} & 58.6 & 72.7 & 87.8 & 90.2 & 52.1 & 81.4 $\pm$ 13.9 \\

  \midrule
  DDA* & 91.4 & 73.9 & 90.8 & 76.9 & 90.9 & 75.8 & 81.9 & 88.6 & 82.4 & 75.2 & 83.6 & 86.2 & 91.8 & 83.8 $\pm$ 6.7 \\

  \addlinespace[1pt]
  \rowcolor[HTML]{E3F1FF}
  \textbf{STAL (ours)} & 97.7 & \textbf{97.9} & 95.9 & 86.5 & 97.8 & \textbf{98.6} & 95.3 & \textbf{97.6} & \underline{93.1} & 81.7 & \underline{95.9} & \underline{96.7} & \textbf{96.4} & \textbf{94.7 $\pm$ 5.0} \\
  \bottomrule
  \end{tabular}%
  }
  \label{tab:forensynths}
\end{appendixtable}

\end{table*}

\FloatBarrier
\end{document}